\documentclass[lettersize,journal]{IEEEtran}
\usepackage{amsmath,amsfonts}
\usepackage{algorithmic}
\usepackage{algorithm}
\usepackage{array}
\usepackage[caption=false,font=normalsize,labelfont=sf,textfont=sf]{subfig}
\usepackage{textcomp}
\usepackage{stfloats}
\usepackage{url}
\usepackage{verbatim}
\usepackage{graphicx}
\usepackage{cite}
\usepackage {amssymb}
\usepackage {pifont}
\usepackage{multirow}
\usepackage{bm}

\usepackage{colortbl}
\allowdisplaybreaks[4]

\begin{document}

\title{Select2Col: Leveraging Spatial-Temporal Importance of Semantic Information for Efficient Collaborative Perception}

\author{Yuntao~Liu,~Qian~Huang,~Rongpeng~Li,~Xianfu~Chen,~Zhifeng~Zhao,~Shuyuan~Zhao,~Yongdong~Zhu and Honggang~Zhang
\thanks{Y. Liu is with the College of Information Science and Electronic Engineering, Zhejiang University, Hangzhou 310027, China as well as with Zhejiang Lab, Hangzhou 311121, China (e-mail: liuyuntao@zju.edu.cn).}
\thanks{Q. Huang, S. Zhao and Y. Zhu are with Zhejiang Lab, Hangzhou 311121, China (e-mail: \{huangq, zhaosy, zhuyd\}@zhejianglab.com).}
\thanks{R. Li is with the College of Information Science and Electronic Engineering, Zhejiang University, Hangzhou 310027, China (e-mail: lirongpeng@zju.edu.cn).}
\thanks{X. Chen is with the VTT Technical Research Centre of Finland, Oulu, 90570, Finland (e-mail: xianfu.chen@vtt.fi).}
\thanks{Z. Zhao and H. Zhang are with Zhejiang Lab, Hangzhou 311121, China (e-mail: \{zhaozf, honggangzhang\}@zhejianglab.com) as well as with the College of Information Science and Electronic Engineering, Zhejiang University, Hangzhou 310027, China.}
}



\maketitle

\begin{abstract}
Collaborative perception by leveraging the shared semantic information plays a crucial role in overcoming the individual limitations of isolated agents. However, existing collaborative perception methods tend to focus solely on the spatial features of semantic information, while neglecting the importance of the temporal dimension. Consequently, the potential benefits of collaboration remain underutilized. In this article, we propose Select2Col, a novel collaborative perception framework that takes into account the \underline{s}patial-t\underline{e}mpora\underline{l} importanc\underline{e} of semanti\underline{c} informa\underline{t}ion. Within the Select2Col, we develop a collaborator selection method that utilizes a lightweight graph neural network (GNN) to estimate the importance of semantic information (IoSI) of each collaborator in enhancing perception performance, thereby identifying contributive collaborators while excluding those that potentially bring negative impact. Moreover, we present a semantic information fusion algorithm called HPHA (historical prior hybrid attention), which integrates multi-scale attention and short-term attention modules to capture the IoSI in feature representation from the spatial and temporal dimensions respectively, and assigns IoSI-consistent weights for efficient fusion of information from selected collaborators. Extensive experiments on three open datasets demonstrate that our proposed Select2Col significantly improves the perception performance compared to state-of-the-art approaches. The code associated with this research is publicly available at https://github.com/huangqzj/Select2Col/.
\end{abstract}


\section{Introduction}
\IEEEPARstart{P}{erception} is an essential capability for agents, especially for autonomous vehicles (AVs). However, in real-world scenarios involving occluded or distant objects, relying solely on single-agent individual perception capabilities often encounters significant challenges \cite{1}, and possibly leads to catastrophic outcomes, as depicted in Fig. 1. Fortunately, the emergence of the Internet of Vehicles (IoV) allows AVs to communicate with other agents, thereby acquiring additional perception information \cite{2}. Consequently, collaborative perception, wherein participating agents (e.g., vehicles and road infrastructures) exchange their perception information to obtain an enhanced understanding of surroundings beyond their individual capabilities, has become a promising solution to avoid perception difficulties encountered by isolated agents\cite{3}.  In an illustrative scenario in Fig. 1, the ego vehicle becomes capable of detecting the occluded Vehicle \textbf{A} by leveraging  perception information provided by other agents (i.e., its collaborators). Hence, it not only enhances the perception performance of the ego vehicle but also improves its ability to respond to abnormal situations, thus mitigating potential traffic conflicts.
\begin{figure}[!t]
\centering
\includegraphics[width=0.48\textwidth]{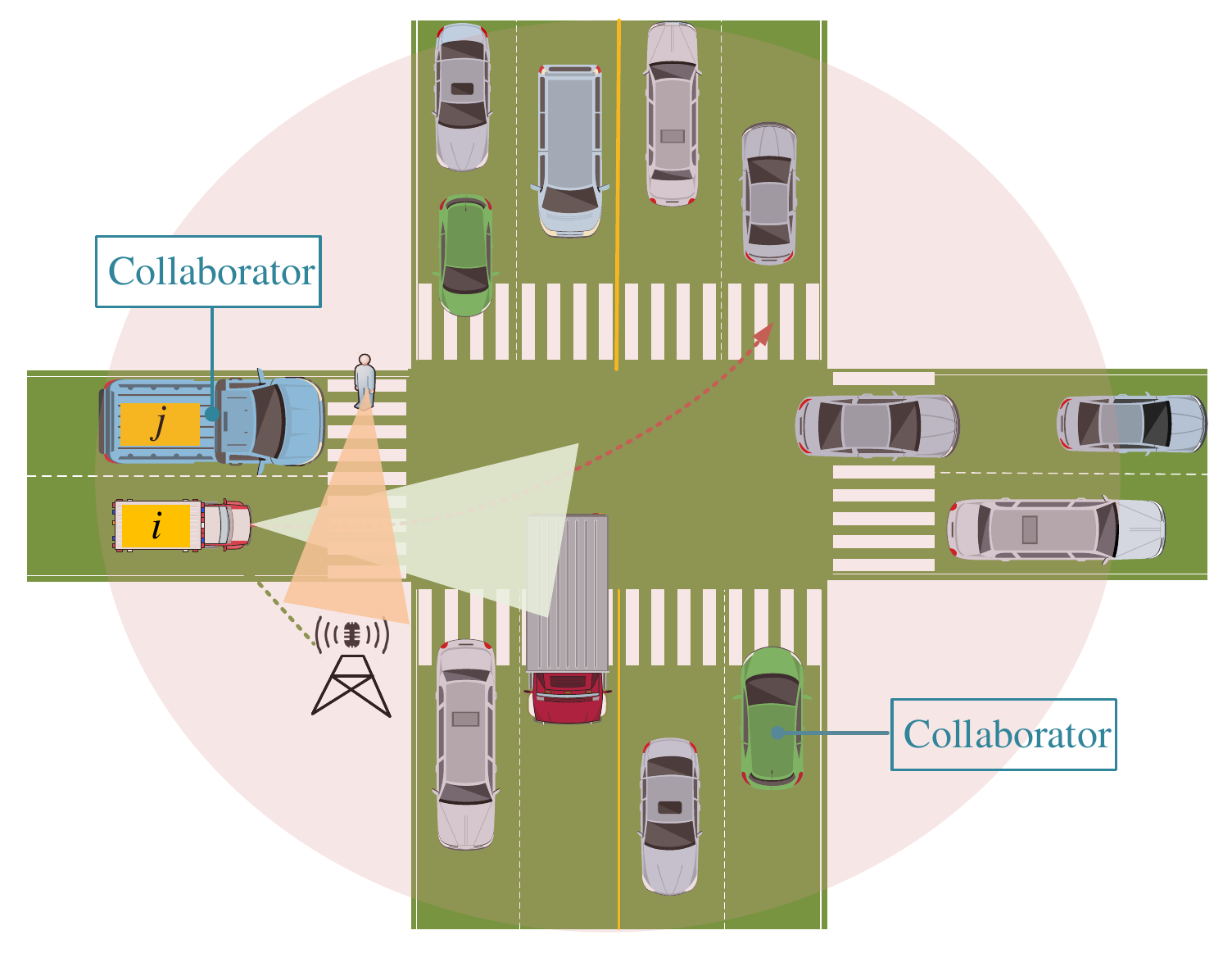}
\caption{The limitation of single-agent perception. From the perspective of the ego vehicle, Vehicle \textbf{A} is occluded, causing a dangerous collision risk when both vehicles accelerate. Such collisions occur frequently.} 
\label{Fig.1}
\end{figure}

In terms of the embedded characteristics of shared information, collaborative perception can be classified as early fusion \cite{4}, late fusion\cite{5,6}, and intermediate fusion\cite{7,8,9,10,11,12,13,14,15,16,17,18}. Specifically, the former two categories, which share the raw data directly or perception results (e.g., object classification and regression results), could either inflict significant communication cost or yield limited performance due to the inadequately shared scene context. Instead, intermediate fusion, which involves to share extracted feature information (i.e., semantic information) from raw perception data, has been consistently demonstrated as a promising strategy to effectively balance perception performance and communication overhead \cite{9,10,11}. 

However, as semantic information suffers from multi-level heterogeneity (e.g., diverse transmission latency, different sensor configurations, and distinct views of agents), the shared semantic information could be rather asynchronous and misaligned in both temporal and spatial dimensions. For instance, the semantic information of different agents may offer inconsistent sizes and locations for the same object. Therefore, the direct fusion of semantic information could be troublesome. Moreover, the selection of appropriate collaborators becomes crucial, as unsuitable candidates have the potential to introduce disruptive noise interference, which ultimately results in a deterioration of the overall perception performance \cite{12}. Nonetheless, identifying beneficial collaborators that can enhance perception performance is not straightforward, due to their diverse spatial-temporal perspectives. 

In order to realize efficient collaborative perception, several recent studies have been proposed \cite{8,9,10,11,12,13,14,15,16,17,18}, and prominent works include CoFF\cite{8}, V2VNet\cite{9}, V2X-Vit\cite{10}, Where2comm\cite{11}, Who2com\cite{12} and When2com\cite{13}. However, these studies solely focus on the spatial aspect of semantic information, disregarding the importance of the temporal dimension. Given the heterogeneous nature of inter-agent sensors and the time required for information transmission, shared semantic information inherently encounters latency in real-world scenarios\cite{10}. Disregarding the temporal dimension of semantic information in collaborative perception results in limited or even degraded performance enhancement. For instance, relying on outdated shared semantic information, such as depicting objects that are no longer present in the focus region, has the potential to confuse the ego and lead to misleading perception outcomes. Therefore, comprehensive consideration of both the spatial and temporal dimensions of semantic information becomes indispensable for successful collaborative perception. Furthermore, considering the diverse temporal and spatial characteristics of semantic information from various collaborators, the importance of semantic information in enhancing perception performance also varies. Therefore, integrating the importance of semantic information (IoSI) into collaborative perception is highly attractive. This can be achieved by assigning distinctive spatial-temporal importance weights to different semantic information and utilizing these weights to guide both collaborator selection and semantic information fusion.

In this article, motivated by the aforementioned discussions and the significant contributions of artificial intelligence studies \cite{19,20,21}, we propose an IoSI-based collaborative perception framework called Select2Col to more comprehensively take account of the \underline{s}patial-t\underline{e}mpora\underline{l} importanc\underline{e} of semanti\underline{c} informa\underline{t}ion. Within the Select2Col framework, we develop an IoSI-based method for collaborator selection, which enables efficient selection of contributive collaborators from both temporal and spatial dimensions. Moreover, we design a historical prior hybrid attention (HPHA) fusion algorithm to achieve more effective information fusion by assigning IoSI-consistent weights. Compared to existing works\cite{8,9,10,11,12,13,14,15,16,17}, the main contributions of this article can be summarized as follows. \par
\begin{itemize}
\item[1)]We propose the Select2Col\footnote{The codes for our proposed Select2Col are available at https://github.com/huangqzj/Select2Col/.} framework for collaborative perception, which allows to capture the IoSI from spatial and temporal dimensions for both effective collaborator selection and meaningful information fusion.
\item[2)]
We design a semantic information fusion algorithm HPHA, which employs a multi-scale attention module and a short-term attention module to efficiently aggregate semantic information from the spatial and temporal dimensions of IoSI. Moreover, we present a collaborator selection method based on IoSI, ensuring that only semantic information from contributive collaborators is leveraged.
\item[3)]
We carry out a comprehensive evaluation of our proposed Select2Col on three open datasets OPV2V\cite{14}, V2XSet\cite{10}, and V2V4Real\cite{22}. The experimental results show that Select2Col is more effective in improving perception performance than state-of-the-art (SOTA) works, such as V2VNet\cite{9}, V2X-Vit\cite{10}, and Where2comm\cite{11}.
\end{itemize}

The remainder of this article is organized as follows. Section II reviews the related works on collaborative perception. Section III presents the system model and formulates the problem. Section IV elaborates on our proposed Select2Col framework. Sections V and VI introduce our innovative collaborator selection method and semantic information fusion algorithm HPHA, respectively. We conduct experiments to verify our proposed Select2Col in Section VII. Finally, we conclude this article with a summary in Section VIII.\par

\section{Related Works}

\subsection{Perception Means in IoV}

Perception plays a crucial role in various tasks of IoV (especially AVs), such as 3D object detection \cite{23}, multiple object tracking\cite{24}, semantic segmentation \cite{25}, and depth estimation \cite{26}. In particular, 3D object detection, which aims to accurately identify predefined object categories by providing classification and regression information \cite{27}, is considered as one of the fundamental perception tasks within the field of computer vision. 

Given commonly utilized perception sensors in autonomous vehicles, methods for 3D target detection can be categorized into camera-based and LiDAR-based methods \cite{28}. Meanwhile, LiDAR-based detection methods are particularly favored due to the inherent 3D information of LiDAR point clouds, which maintain excellent performance even under challenging weather and lighting conditions \cite{27}. However, LiDAR point clouds often suffer from sparsity and disorderliness, which pose challenges for the direct processing of individual points  \cite{29}. To address this limitation, it commonly turns point clouds into alternative structures, such as voxel \cite{30} or pillar \cite{31}, to improve the processing efficiency. Notably, compared to voxels, pillar structures offer advantages in terms of reduced computational and memory resources\cite{32}. For example, PointPillars \cite{33}, a widely adopted technique for object detection in point clouds, organizes point clouds into vertical columns and operates on pillars instead of voxels. Consequently, all critical operations can be performed efficiently using 2D convolutions, thus eliminating the need for manual tuning of vertical direction binning. Given the requirement for real-time inference in detection models for autonomous vehicles, many state-of-the-art collaborative perception methods, such as V2VNet\cite{9}, V2X-Vit\cite{10}, and Where2comm\cite{11}, integrate the backbone of PointPillars in their respective models.

Consistent with these works, we primarily consider  LiDAR-based 3D object detection as our perception task and focus on how to enhance the performance of collaborative perception, given the  inherent limitations of single-agent perception \cite{32}.\par

\begin{table}[!t]
\centering
\caption{A comparison between Select2Col and highly related works.}
\renewcommand{\arraystretch}{1.2}
\begin{tabular}{|m{2.2cm}<{\centering}|m{2.2cm}<{\centering}|m{2.2cm}<{\centering}|}
\hline
\textbf{Works} & \textbf{Collaborator selection strategy} & \textbf{Information fusion strategy} \\ \hline
\cite{8,9,10}          &All agents                                                                             &Spatial fusion                                                              \\ \hline
\cite{15,16,17,18}       &All agents        &Spatial fusion \\ \hline

\cite{11,12,13}     &Spatial selection                                                        &Spatial fusion                                                              \\ \hline
\textbf{Select2Col (Ours)}        &\textbf{Spatial-temporal selection}            &\textbf{Spatial-temporal fusion}                                                    \\ \hline
\end{tabular}
\end{table}

\subsection{Related Studies in Collaborative Perception}

Collaborative perception allows agents to share the perception information through the IoV, overcoming the inherent limitations of single-agent perception. Early research primarily focuses on effectively fusing feature information from collaborators, with the pioneering V2VNet \cite{9} being one such  example. Specifically, V2VNet employs a space-aware graph neural network (GNN) to aggregate information from neighboring agents. Similarly, in \cite{15}, a collaborative perception solution is proposed by using a graph attention network-based aggregation strategy to fuse intermediate representations. In addition, V2X-ViT \cite{10} introduces a unified vision transformer architecture with heterogeneous multi-agent self-attention to effectively capture the spatial relationship between agents, leading to efficient feature fusion. Another related work \cite{16} designs adaptive feature fusion models using trainable neural networks in the spatial dimension. Meanwhile, the CoFF approach \cite{8} fuses feature information by enhancing the spatial feature information of the original perception data. In another related study \cite{17}, a feature fusion method that leverages a repair network module and a specially designed V2V attention module is proposed to mitigate the influence of shared feature loss during transmission. Furthermore, CoAlign \cite{18} introduces a collaborative perception framework that effectively handles unknown pose errors by utilizing a novel agent-object pose graph model.

Although the incorporation of multiple agents in collaborative perception yields performance improvement, studies of Who2com\cite{12} and literature \cite{34} demonstrate that bluntly adding more collaborators does not always guarantee performance benefits and can even result in counter-intuition negative effects. Therefore, the selection of effective collaborators and the fusion of meaningful information becomes crucial for collaborative perception. For example, When2com\cite{13} proposes to forge several communication groups from neighboring agents in terms of the spatial feature similarity of the shared information and determines when to collaborate within these groups. On the other hand, Where2comm\cite{11} employs a spatial confidence map to select collaborators and utilizes multi-head attention to fuse the shared semantic information. 

Despite the remarkable progress, the lack of exploring the temporal dimension of semantic information makes these works far from optimality. To address this performance gap, our proposed Select2Col highlights the importance of both spatial and temporal dimensions of IoSI for  collaborator selection and semantic information fusion, thereby achieving further improvements in perception performance. As outlined in TABLE I, Select2Col exhibits notable distinctions from existing works.\par

\section{System Model and Problem Formulation}

\subsection{System Model}
Beforehand, Table II presents a list of notations used in this manuscript.\par

Consistent with SOTA methods, such as\cite{9,10,11}, we adopt a collaborative perception scenario wherein at a given time-slot $t$, there exists an ego agent \emph{i} performing collaborative perception with a set  of  neighboring agents ${\mathcal{N}_{i}^{t}}$ (e.g., vehicles and/or road infrastructures with perception and communication functionalities), as depicted in Fig. 1. From the ego's perspective, the objective of collaborative perception is to effectively leverage the received  semantic information to maximize its perception performance.\par

Mathematically, we denote that at the very beginning of $t$, agent $i$ obtains its raw perception data $X_{i}$ from its sensors and receives semantic information transmitted by its neighboring agents. From that time, the collaborative perception undergoes the following procedures.\par

\begin{table}[]
\centering
\caption{Notations and Explanation}
\renewcommand{\arraystretch}{1.15}
\begin{tabular}{|c|l|}
\hline
\multicolumn{1}{|c|}{\textbf{Notation}} &  \multicolumn{1}{c|}{\textbf{Explanation}}  
 \\ \hline
$X_{i}$       & Raw perception data of agent \emph{i}                      \\
\hline
${\mathcal{N}_{i}^{t}}$       & Set of neighbor agents for agent \emph{i} at time-slot $t$                   \\
\hline
${\mathcal{C}_{i}^{t}}$       & Set of selected collaborators for agent \emph{i} at time-slot $t$                            \\
\hline
${\mathcal{S}}$       & Set of  scales  for mult-scale attention operation                      \\
\hline

${\bm{F}_{i}^{t}}$         & \begin{tabular}[c]{@{}l@{}} Semantic information of agent \emph{i} at time-slot $t$\end{tabular}                                       \\
\hline

${\bm{F}_{j}^{ t-\tau _{ji}}}$  & \begin{tabular}[c]{@{}l@{}}Semantic information of agent \emph{j}  shared to agent \emph{i} \\ at time-slot \emph{t} \end{tabular} \\
\hline

$\bm{H}_{s}$  & \begin{tabular}[c]{@{}l@{}}Aggregated feature semantic information after   \\spatial attention operation at a specific scale $s$\end{tabular}                                                \\ 
\hline
$\bm{H}_\mathrm{ms}$  & \begin{tabular}[c]{@{}l@{}}Aggregated feature semantic information after  \\ multi-scale attention operation\end{tabular}        \\    
\hline

$\bm{H}_{i}$       & \begin{tabular}[c]{@{}l@{}}Fused semantic information of agent \emph{i} \\  \end{tabular}                                                 \\        
\hline  

$\bm{M}_j$       & \begin{tabular}[c]{@{}l@{}}Sparse map of agent \emph{j} \\\end{tabular}                                                 \\     
\hline

$\bm{W}_{s,ji}^\mathrm{{sa}}$ & \begin{tabular}[c]{@{}l@{}}Spatial-attention weight of agent \emph{j} to agent \emph{i} at  \\ scale $s$ \end{tabular}                                                \\
\hline

$\bm{W}^\mathrm{{ta}}$ &Temporal-attention weight \\
\hline

$w_{ji}^\mathrm{{en}}$       & \begin{tabular}[c]{@{}l@{}}Enhanced weight of agent \emph{j} to agent \emph{i} \\ \end{tabular}                                                 \\    
\hline

$\widehat{Y}_{i}$       & \begin{tabular}[c]{@{}l@{}}Perception results of agent \emph{i} \end{tabular}                                          \\
\hline
$Y_{i}$       & \begin{tabular}[c]{@{}l@{}}Ground truth of the perception results of  agent \emph{i} \\ \end{tabular}                                                 \\
\hline

$f_\mathrm{{enc}}$   & \begin{tabular}[c]{@{}l@{}}Encoder that extracts  semantic information from raw\\    perception data\end{tabular}                                         \\
\hline
$f_\mathrm{{select}}$   & \begin{tabular}[c]{@{}l@{}}Selector that outputs the selected collaborators\\ \end{tabular}                                         \\
\hline
$f_\mathrm{{fuse}}$   & \begin{tabular}[c]{@{}l@{}}Fuser that outputs the fused semantic information \\ \end{tabular}                                         \\
\hline
$f_\mathrm{{dec}}$   & \begin{tabular}[c]{@{}l@{}}Decoder that outputs perception results  \\  \end{tabular}                                      \\
\hline
$f_\mathrm{{eva}}$   & \begin{tabular}[c]{@{}l@{}}Standard perception performance evaluation function  \\  \end{tabular}                                      \\
\hline

$\mathrm{conv}\left( \cdot \right)$   & \begin{tabular}[c]{@{}l@{}}Convolution block operation  \\  \end{tabular}                                      \\
\hline

$\mathrm{deconv}\left( \cdot \right)$    & \begin{tabular}[c]{@{}l@{}}Deconvolution block operation \\  \end{tabular}                                      \\
\hline

$\mathrm{sa}\left( \cdot \right)$    & \begin{tabular}[c]{@{}l@{}}Spatial attention operation   \\  \end{tabular}                                      \\
\hline

$\mathrm{ta}\left( \cdot \right)$    & \begin{tabular}[c]{@{}l@{}}Temporal attention operation  \\  \end{tabular}                                      \\
\hline

$\tau _{ji}$  &Information availability latency of agent \emph{j} to agent \emph{i} \\

\hline
$t_{ji}^{\mathrm{asyn}}$   & \begin{tabular}[c]{@{}l@{}}Asynchronous overhead between the sensors of   \\ agent \emph{j} and agent \emph{i} \end{tabular}                      \\
\hline
$t_{ji}^\mathrm{{ext}}$       & \begin{tabular}[c]{@{}l@{}}Semantic information extraction latency of agent \emph{j}  \end{tabular} \\          
\hline                                         
$t_{ji}^\mathrm{{tx}}$      & \begin{tabular}[c]{@{}l@{}}Transmission latency of shared semantic information \\ from  agent \emph{j} to agent \emph{i} \end{tabular}                               \\
\hline
$t_{ji}^\mathrm{{idle}}$      & \begin{tabular}[c]{@{}l@{}}Idle time between  the perception system and  \\   communication system \\   \end{tabular}                               \\
\hline
$b_{ji}$       & Bandwidth of agent \emph{j} to agent \emph{i}                                                                                                                               \\
\hline
$p_{ji}^\mathrm{{tx}}$      & \begin{tabular}[c]{@{}l@{}}Transmission power of agent \emph{j} to agent \emph{i} \\ \end{tabular} \\    
\hline
$p_{ji}^\mathrm{{loss}}$      & Path loss between agent \emph{j} and agent \emph{i}                                                                                                               \\
\hline

$p_{ji}^\mathrm{{noise}}$      & \begin{tabular}[c]{@{}l@{}}Noise power between agent \emph{j} and agent \emph{i} \\  \end{tabular} \\    
\hline
$d_{ji}$     & Distance between agent \emph{j} and \emph{i}                                                                                                                 \\
\hline
$f_\mathrm{c}$       & Center frequency in GHz                                                                                                                                        \\
\hline
$\mathrm{s}_{\mathrm{ms}}$    &  Multi-scale attention output  scale                                                                     \\
\hline
 $T$    &  Sensor sampling interval                                                                                                                \\
\hline
 $K$    &  Number of historical prior frames                                                                                                                 \\
\hline

\end{tabular}
\end{table}

\subsubsection{Semantic information extraction}
We adopt the intermediate feature fusion strategy\cite{9}. In other words, et each time-slot \emph{t}, agent \emph{i} employs an encoder $f_{\mathrm{enc}}$ to extract semantic information from its raw perception data $X_{i}$, such as LiDAR point cloud data \cite{11}. For any agent $i$, the corresponding semantic information $\bm{F}_{i}^{ t }$ is extracted at time-slot $t$ as follows
\begin{align}
\bm{F}_{i}^{t}=f_{\mathrm{enc}}\left( X_{i} \right). 
\end{align}
In addition, for a simpler and more consistent representation of semantic information, we employ the notation $\bm{F}_{j}^{t-\tau _{ji}}$  to denote the semantic information of agent $j\in \mathcal{N}_{i}^{t}\cup \left\{ i \right\}$, where  $\tau _{ji}$ represents the information  latency between the source agent \emph{j} and the target agent \emph{i}. Specifically, when \emph{j} equals \emph{i}, $\tau _{ji}$ is set to 0 as there is no latency between agent \emph{i} and itself. Otherwise, $\tau _{ji}$ can be defined as the sum of inter-agent asynchronous overhead $t_{ji}^{\mathrm{asyn}}$, the semantic information extraction latency $t_{ji}^{\mathrm{ext}}$, the semantic information transmission latency $t_{ji}^{\mathrm{tx}}$, and the idle time $t_{ji}^{\mathrm{idle}}$ caused by the lack of synchronization between the perception system and communication system\cite{10}. In other words,
\begin{align}
\tau _{ji}=t_{ji}^{\mathrm{asyn}}+t_{ji}^{\mathrm{ext}}+t_{ji}^{\mathrm{tx}}+t_{ji}^{\mathrm{idle}}.
\end{align}
According to the specification 3GPP TR 38.901 \cite{35}, network transmission latency $t_{ji}^{\mathrm{tx}}$ can be approximately derived from a path-loss driven channel as
\begin{equation}
\begin{aligned}
t_{ji}^{\mathrm{tx}}\,\,=\,\,\frac{\mathrm{size}\left(\bm{F}_{j}^{t-\tau _{ji}} \right)}{b_{ji}\log _2\left.( 1+10^{0.1\left(p_{ji}^{\mathrm{tx}}\,\,-p_{ji}^{\mathrm{loss}}-p_{ji}^{\mathrm{noise}} \right)} \right.)},
\end{aligned}
\end{equation}
where $\mathrm{size}\left( \cdot \right) $ denotes a function that calculates the size of the transmitted semantic information. Besides, $b_{ji}$, $p_{ji}^{\mathrm{tx}}$, $p_{ji}^{\mathrm{noise}}$ and $p_{ji}^{\mathrm{loss}}=28.0+22\log _{10}\left( d_{ji} \right) +20\log _{10}\left( f_{\mathrm{c}} \right)$ represent agent \emph{j}'s transmission bandwidth, transmission power, transmission noise power, and  transmission path loss to agent \emph{i}, respectively. Furthermore, $d_{ji}$ denotes the distance between agent \emph{j} and  \emph{i} in meters, and  $f_{\mathrm{c}}$ represents  the center frequency in GHz. \par

\subsubsection{Contributive collaborator selection}
The semantic information of neighboring agents is derived from a variety of raw perception data with various spatial and temporal dimensions.  It is important to note that not all semantic information is beneficial. In certain cases, the semantic information provided by specific agents may degrade perception performance due to excessive delay or unreasonable spatial viewpoint \cite{12}. Therefore, agent \emph{i} employs  a selector $f_{\mathrm{select}}$ to determine a contributive collaborator set $\mathcal{C}_{i}^{t }$ from  $\mathcal{N}_{i}^{t }$, namely,
\begin{align}
\mathcal{C}_{i}^{t}=f_{\mathrm{select}}\Big(\big\{ \bm{F}_{j}^{t-\tau _{ji}} \big\} _{j\in \mathcal{N}_{i}^{t} \cup \left\{ i \right\} } \Big).
\end{align}

\subsubsection{Semantic information fusion}
After selecting contributive collaborators, agent \emph{i} utilizes a fuser $f_{\mathrm{fuse}}$  to aggregate its semantic information $\bm{F}_{i}^{t}$  with selected collaborators' semantic information $\left.\{ \bm{F}_{j}^{t-\tau _{ji}} \right.\} _{j\in \mathcal{C}_{i}^{t}}$  and correspondingly obtain the fused semantic information $\bm{H}_{i}$ with a more comprehensive representation of features.  The semantic information fusion expression can be written as 
\begin{align}
\bm{H}_{i}=f_{\mathrm{fuse}}\Big( \big\{ \bm{F}_{j}^{t-\tau _{ji}} \big\} _{j\in \mathcal{C}_{i}^{t} \cup \left\{ i \right\} } \Big).
\end{align}

\subsubsection{Semantic information decoding}
Finally, for each agent $i$, the fused semantic information $\bm{H}_{i}$ needs to be converted into appropriate perception results $\widehat{Y}_{i}$ by a decoder $f_{\mathrm{dec}}$. Mathematically, the expression for this process is as 
\begin{align}
\widehat{Y_{i}}=f_{\mathrm{dec}}(\bm{H}_{i}).
\end{align}

\subsection{Problem Formulation}
Consistent with all collaborative perception works, we use the average precision (AP) metric to evaluate the perception performance, that is
\begin{align}
\mathrm{AP}=\,\,f_{\mathrm{eva}}(\widehat{Y_{i}},Y_{i}),
\end{align}
where $Y_{i}$ represents the ground truth of agent \emph{i}'s perception results, and $f_{\mathrm{eva}}\left( \cdot \right) $ is a standard perception performance evaluation function \cite{36} that measures the accuracy of the perception results relative to the ground truth. 

Thus, in order to maximize the perception performance (i.e., maximize the AP), together with (4) to (7), the  collaborative perception problem can be formulated as 

\begin{equation}
\begin{aligned}
\max \mathrm{AP}=f_{\mathrm{eva}}\Bigg( f_{\mathrm{dec}}\Big( f_{\mathrm{fuse}}\big( \big\{ \bm{F}_{j}^{t-\tau _{ji}} \big\} _{j\in \mathcal{C}_{i}^{t} \cup \left\{ i \right\}      } \big) \Big) , Y_{i} \Bigg) \\
 s.t. \ \mathcal{C}_{i}^{t}=f_{\mathrm{select}}\Big( \big\{ \bm{F}_{j}^{t-\tau _{ji}} \big\} _{j\in \mathcal{N}_{i}^{t} \cup \left\{ i \right\}  } \Big) .\hspace{0.8cm} 
\end{aligned}
\end{equation}

As implied in (8), it turns to optimize the design of the selector $f_{\mathrm{select}}\left( \cdot \right)$ and fuser $f_{\mathrm{fuse}}\left( \cdot \right)$, since the $f_{\mathrm{enc}}\left( \cdot \right) $,  $f_{\mathrm{dec}}\left( \cdot \right)$, and  $f_{\mathrm{eva}}\left( \cdot \right)$  are typically common and well-established elements\cite{9,10,11}. Furthermore, the former $f_{\mathrm{select}}\left( \cdot \right)$ involves identifying effective collaborators that can provide high-quality perception semantic information, while the latter considers the correlation of semantic information to improve the feature representations. Nevertheless, it is non-trivial to describe these two functions in compact, mathematical formulations. Hence, vanilla approaches \cite{9,10,11,12,13,14} approximate these functions with the guidance of intuitive observations. For example, collaborators with larger ego interest regions promise to be more contributive in improving the perception performance of ego. Furthermore, assigning higher weights to more important semantic information for fusion can contribute to yielding an enhanced feature representation. However, these approaches ignore the importance of semantic information from the temporal dimension, resulting in limitations in accurately fitting these functions.\par

Therefore, to address these challenges, we leverage lightweight GNN and hybrid attention modules to calibrate the selector  $f_{\mathrm{select}}\left( \cdot \right)$ and fuser  $f_{\mathrm{fuse}}\left( \cdot \right)$ based on the IoSI from spatial-temporal dimensions. The details shall be given in Section V and Section VI. \par

\begin{figure*}[]
\centering
\includegraphics[width=1\textwidth]{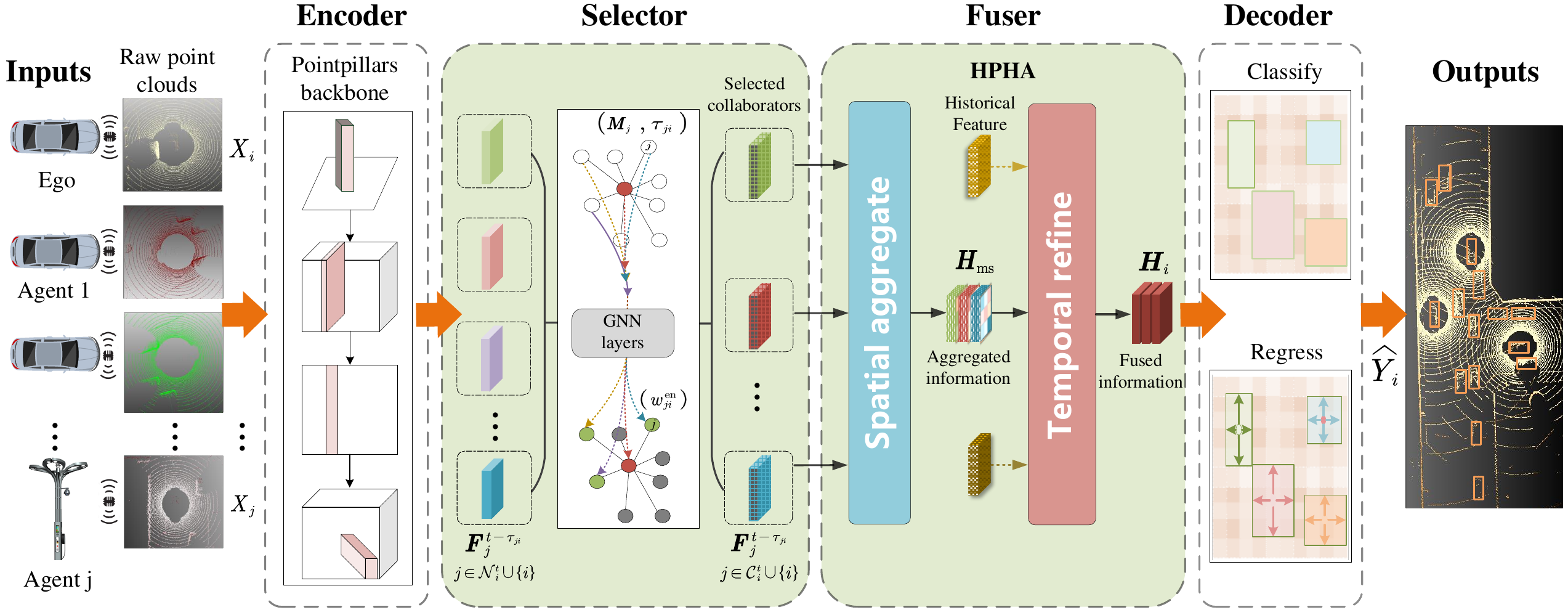}
\caption{The overview of our proposed Select2Col framework.} 
\label{Fig.2}
\end{figure*}

\section{Select2Col: An IoSI-based Collaborative Perception Framework}
Based on the previous analysis, we propose Select2Col, an IoSI-based collaborative perception framework. As depicted in Fig. 2, Select2Col comprises four components (i.e., encoder $f_{\mathrm{enc}}$, selector $f_{\mathrm{select}}$, fuser $f_{\mathrm{fuse}}$, and decoder  $f_{\mathrm{dec}}$) to implement the entire process of collaborative perception. We briefly introduce these components of Select2Col in this section, while leaving the innovative design of collaborator selection and information fusion in Section V and Section VI, respectively.\par

\subsection{Overview of Select2Col}
Typically, the Select2Col framework encompasses the following components.\par

\subsubsection{Encoder}The encoder extracts semantic information from raw perception data. In this article, we focus on LiDAR-based 3D object detection as the perception task. Correspondingly, we adopt the backbone of Pointpillars\cite{33} as our encoder because of its low inference latency and optimized memory usage, consistent with the literature\cite{9,10,11}.\par

\subsubsection{Selector} This selector is invoked to determine contributive collaborators $\mathcal{C}_{i}^{t}$ using the semantic information $\left.\{ \bm{F}_{j}^{t-\tau _{ji}} \right.\} _{j\in \mathcal{N}_{i}^{ t } \cup \left\{ i \right\}}$ obtained from the outputs of the encoder. As mentioned earlier, we design an IoSI-based collaborator selection method, the details of which are given in Section V.\par

\subsubsection{Fuser}This fuser is responsible for obtaining the fused semantic information $\bm{H}_{i}$ using $\left.\{\bm{F}_{j}^{t-\tau _{ji}} \right.\} _{j\in \mathcal{C}_{i}^{t}\cup \left\{ i \right\}}$. In particular, we propose a semantic information fusion algorithm named HPHA to enhance the fusion feature representation as further discussed in Section VI.\par

\subsubsection{Decoder}This decoder outputs the perception results $\widehat{Y}_{i}$ based on the fused semantic information $\bm{H}_{i}$. Consistent with \cite{11}, we utilize a two-layer convolution block to downsample the fused semantic information and then employ a single-layer convolutional classification head and a single-layer convolutional regression head to obtain the classification and regression information of objects, respectively. \par

\subsection{End-to-End Network Model}
We employ an end-to-end neural network model to effectively integrate the components above, given that the output of each component serves as the input for the next one.\par

This network model takes the raw perception data of the ego agent and its neighboring agents as input. It produces collaborative perception results (i.e., the results of 3D object detection, including the classification and regression information) for the ego agent. Therefore, the loss function of the model consists of a classification loss and a regression loss. In line with PointPillars\cite{33}, we utilize focal loss \cite{37} for the classification loss  and smooth L1 loss \cite{38} for the regression loss. \par
Furthermore, the network model is trained in an end-to-end manner under the supervision LiDAR-based 3D object detection task. During training, a random agent is selected as the ego from the training subset of the open datasets, such as OPV2V\cite{14}, V2XSet\cite{10}, and V2V4Real\cite{22}. The training method and parameters are consistent with the SOTA works\cite{9,10,11}, and more details can be found in our open-source code.  \par

\begin{figure}[!t]
\centering
\includegraphics[width=0.48\textwidth]{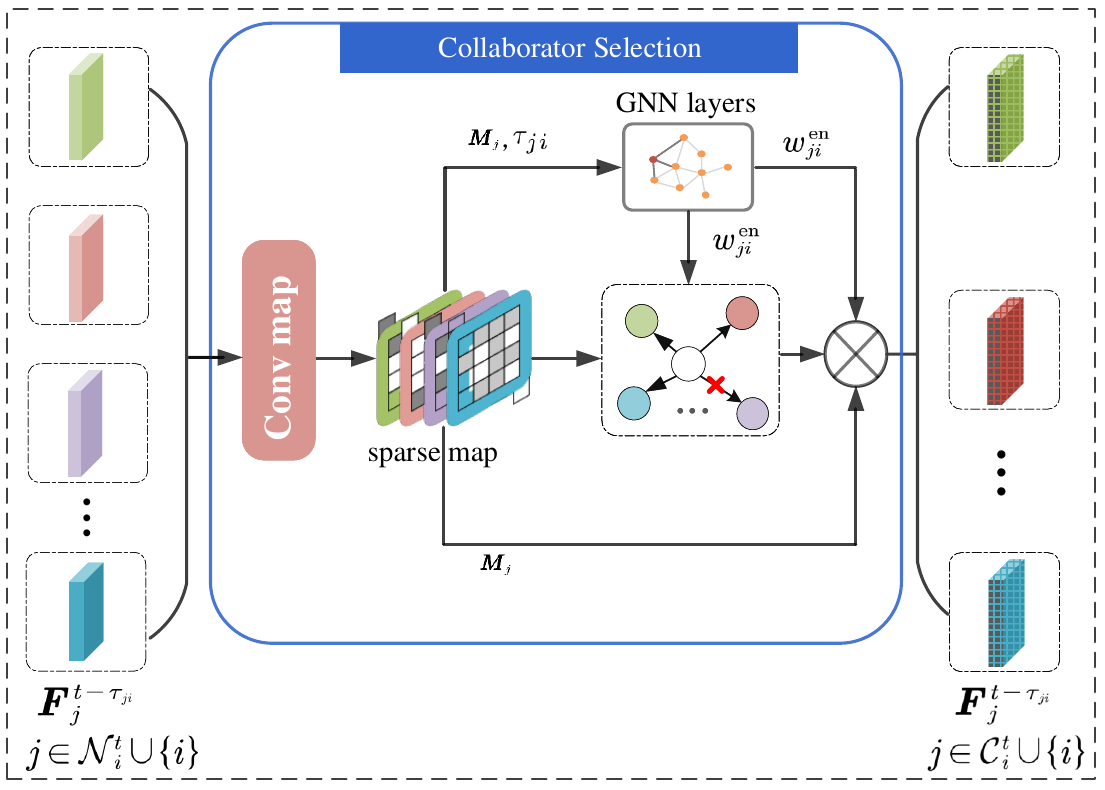}
\caption{Diagram of our collaborator selection method.} 
\label{Fig.3}
\end{figure}

\section{Collaborator Selection Based on IoSI}
Given the spatial-temporal variability of agents, this section introduces our proposed IoSI-based collaborator selection method, that is the fuser $f_{\mathrm{fuse}}\left( \cdot \right)$, to address the challenges of identifying contributive collaborators.

Specifically, the collaborator selection method, as illustrated in Fig. 3, consists of two steps. In the first step, we employ a lightweight GNN module to estimate the enhanced weight of each agent from spatial-temporal dimensions. These enhanced weights indicate the individual IoSI in enhancing the perception performance of the ego. In the subsequent step, contributive collaborators are selected based on the acquired enhanced weights, and the feature information of the chosen collaborators is enhanced accordingly. Different from solely relying on spatial relationships to determine contributive collaborators, our method takes into account both spatial and temporal dimensions, resulting in a more precise selection of contributors and more effective utilization of their feature information.

\subsection{Enhanced Weight Estimation}
For each agent \emph{j} in $\mathcal{N}_{i}^{ t }\cup \left\{ i \right\}$, we first utilize a convolution module $\mathrm{conv}_{\mathrm{map}}$ \cite{11} to extract its  sparse map $\bm{M}_j$ from its semantic information $\bm{F}_{j}^{ t-\tau _{ji}}$. Mathematically, the expression for this process is as
\begin{align}
\bm{M}_j=\mathrm{conv}_{\mathrm{map}}\big( \bm{F}_{j}^{ t-\tau _{ji}} \big)\label{convmap}.
\end{align}

In addition, given that the sparse map contains the spatial location information of potential objects, the element in the sparse map takes either $0$ or $1$ depending on the existence of potential objects (i.e., $1$ indicates the existence, while $0$ implies only the existence of the background).

Subsequently, we employ a lightweight GNN\cite{19} as the backbone to estimate the importance of each agent in enhancing ego \emph{i}'s perception performance from spatial-temporal dimensions. Within this GNN module, each node corresponds to an agent and comprises its sparse map along with the semantic information latency, which respectively indicates the spatial and temporal characteristics of this agent. For each agent \emph{j} in  $\mathcal{N}_{i}^{ t }$, the trained GNN generates its enhanced weight $w_{ji}^{\mathrm{en}}$, reflecting the importance of this agent's semantic information to the ego.

\begin{algorithm}[!t]
\caption{Collaborator Selection Method}\label{alg:alg1}
\begin{algorithmic}[1]
\STATE {\textbf{Input}\textsc{:}}  semantic information of ego and its neighboring agents  $\left.\{ \bm{F}_{j}^{t-\tau _{ji}} \right.\} _{j\in \mathcal{N}_{i}^{ t } \cup \left\{ i \right\}}$.
\STATE {\textbf{Output}\textsc{:}}  selected collaborator set $\mathcal{C}_{i}^{t}$, and corresponding enhanced semantic information $\left.\{ \bm{F}_{j}^{t-\tau _{ji}} \right.\} _{j\in \mathcal{C}_{i}^{t}\cup \left\{ i \right\}}$.
\STATE initialize $\mathcal{C}_{i}^{ t}$ as neighboring agent set $\mathcal{N}_{i}^{ t }$. 
\STATE  \textbf{for} each agent $j\in \mathcal{N}_{i}^{ t}\cup \left\{ i \right\} $  \textbf{do} 
\STATE \hspace{0.5cm} get the corresponding sparse map  $\bm{M}_j$ using (9); 
\STATE   \hspace{0.5cm} \textbf{if} $\left( j=i \right)$  \textbf{then} 
\STATE \hspace{1cm}the latency between ego and itself is 0, i.e., $\tau _{ji}=0$;
 \STATE \hspace{0.5cm}  \textbf{endif}
\STATE \hspace{0.5cm} using $\left( \bm{M}_j,\tau _{ji} \right)$ as agent's spatial-temporal information;
 \STATE  \textbf{endfor}
\STATE agents in $ \mathcal{N}_{i}^{ t}\cup \left\{ i \right\} $ as nodes to build the GNN network;
\STATE GNN outputs each agent's enhanced weight $\left\{ w_{ji}^{en} \right\} _{j\in \mathcal{N} _{i}^{t}}$;

\STATE  \textbf{for} each agent $j\in \mathcal{N}_{i}^{ t}$  \textbf{do} 
\STATE \hspace{0.5cm}  \textbf{if} $\left( w_{ji}^{en}\leqslant w_{threshold} \right) $ \textbf{then} 
\STATE \hspace{1cm} delete invalid agent $j$ from  $\mathcal{C}_{i}^{t }$;
\STATE \hspace{0.5cm}  \textbf{else then} 
\STATE \hspace{1cm} enhanced semantic information of selected collaborator using (10);
\STATE \hspace{0.5cm}  \textbf{endif} 
\STATE \textbf{endfor} 
\STATE \textbf{Return} $\mathcal{C}_{i}^{ t }$, $\left.\{ \bm{F}_{j}^{t-\tau _{ji}} \right.\} _{j\in \mathcal{C}_{i}^{t}\cup \left\{ i \right\}}$.
\end{algorithmic}
\label{alg1}
\end{algorithm}

\subsection{Collaborator Selection and Enhancement}
 We utilize the acquired enhanced weights to determine the contributive agents. For any agent \emph{j} in $\mathcal{N}_{i}^{ t }$, if its enhanced weight $w_{ji}^{\mathrm{en}}$ exceeds a pre-defined threshold, we consider it as a contributive agent; otherwise, we classify it as an ineffective collaborator. After eliminating all ineffective collaborators in $\mathcal{N}_{i}^{ t }$, we then obtain a selected collaborator set $\mathcal{C}_{i}^{ t }$. \par

It is important to note that not all selected individuals contribute equally to the improvement of perception performance. Inspired by the idea in \cite{8}, we employ the acquired enhanced weights to appropriately enhance the features of the selected collaborators. Specifically, for each selected collaborator, such as agent \emph{j}, we perform multiplication between its original feature information $\bm{F}_{j}^{ t-\tau _{ji}}$  and the respective enhanced weights $w_{ji}^{\mathrm{en}}$. Then, we proceed to conduct elementwise multiplication of the enhanced feature information with its sparse graph $\bm{M}_j$. The enhanced semantic information $\bm{F}_{j}^{t-\tau _{ji}}$ for agent \emph{j} in  $ \mathcal{C}_{i}^{ t }$ is obtained as 
\begin{align}
\bm{F}_{j}^{t-\tau_{ji}}=\bm{F}_{j}^{ t-\tau _{ji}} \times w_{ji}^{\mathrm{en}} \otimes \bm{M}_j,
\end{align}
where $\otimes$ represents the elementwise multiplication. Recalling the definition of sparse map $\bm{M}_j$ in (9), (10) implies the enhanced weight $w_{ji}^{\mathrm{en}}$ is only applied to regions with potential objects, which can effectively improve the object features while suppressing background noise.  \par

In conclusion, by identifying contributive collaborators and enhancing their semantic information based on IoSI, the collaborator selection method provides high-quality perception information for subsequent semantic information fusion, thereby further ensuring the effectiveness of the fused semantic information. We summarize the details of the collaborator selection method in Algorithm 1. \par

\begin{figure}[!t]
\centering
\includegraphics[width=0.48\textwidth]{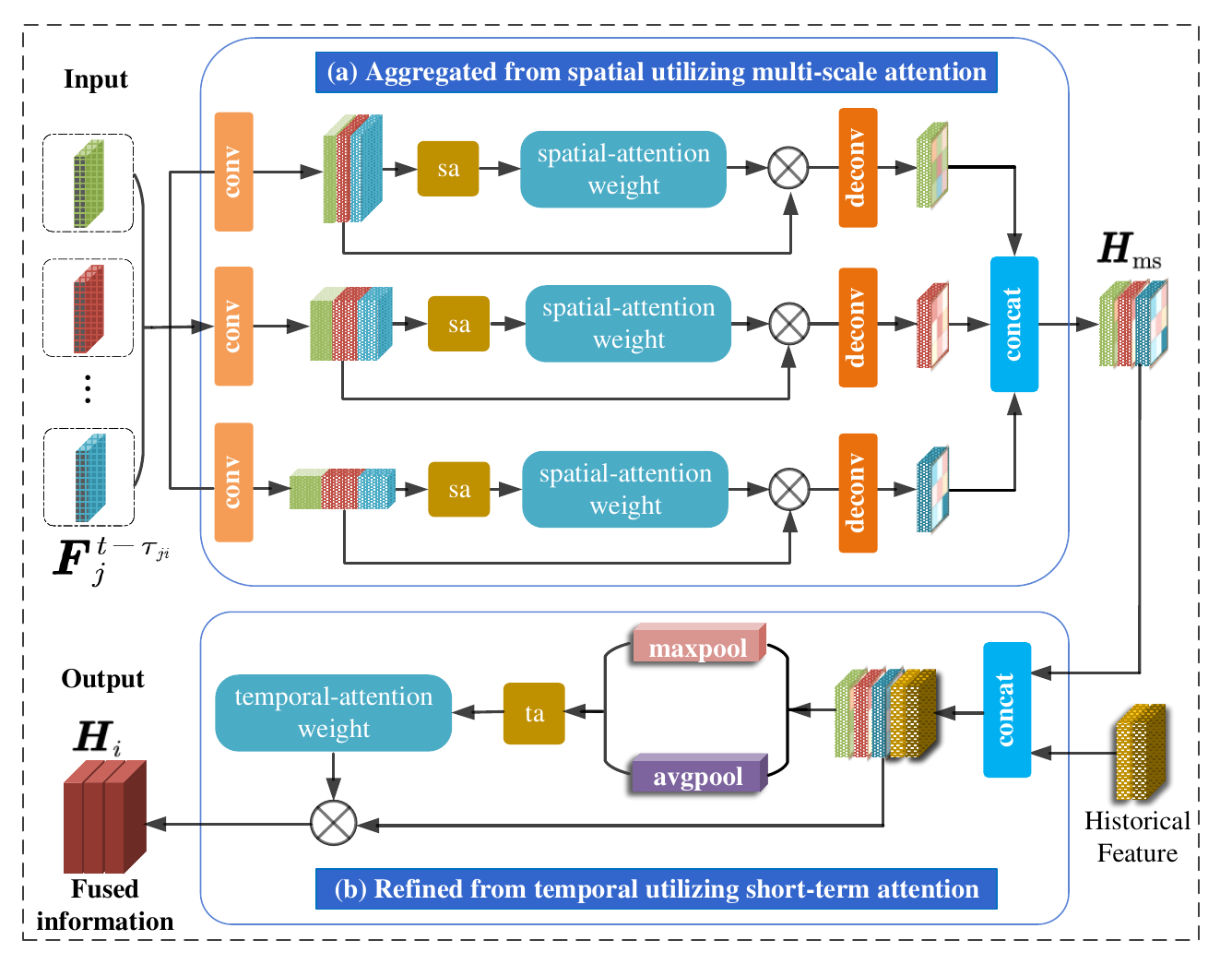}
\caption{Diagram of our HPHA fusion algorithm.} 
\label{Fig.4}
\end{figure}

\section{HPHA: A Historical Prior Hybrid Attention Information Fusion algorithm}

In this section, we introduce our proposed semantic information fusion algorithm HPHA to develop the fuser $f_{\mathrm{fuse}}\left( \cdot \right)$.

As illustrated in Fig. 4, to enhance the feature presentation of the aggregated semantic information, HPHA leverages a multi-scale attention module and a short-term attention module to learn the IoSI in feature presentation from spatial and temporal dimensions respectively, and assigns IoSI-consistent weights to different semantic information for optimal feature fusion. Specifically, the multi-scale attention module captures the correlation between the ego and its collaborators under various  spatial resolutions, while the short-term attention module effectively measures the temporal importance in terms of the correlation between past and present features. Furthermore, our proposed HPHA algorithm is outlined in Algorithm 2.\par

\begin{algorithm}[!t]
\caption{HPHA: Information Fusion Algorithm}\label{alg:alg2}
\begin{algorithmic}[1]
\STATE {\textbf{Input}\textsc{:}} semantic information of ego and its selected collaborators  $\left.\{ \bm{F}_{i}^{t-k\mathrm{T}} \right\} _{k=\left\{0,1\cdots \mathrm{K} \right.\}}$, $\left.\{ \bm{F}_{j}^{t-\tau _{ji}} \right.\} _{j\in \mathcal{C}_{i}^{t}}$.

\STATE {\textbf{Output}\textsc{:}}    fused semantic information  $\bm{H}_{i}$.
\STATE  \textbf{for}   each scale $s$ in $\mathcal{S}$ \textbf{do} 
\STATE \hspace{0.5cm} \textbf{for} each agent $j\in C_{i}^{ t }\cup \left\{ i \right\} $ \textbf{do} 
\STATE \hspace{1cm} calculate the corresponding spatial-attention weight $\bm{W}_{\!s,ji}^{\mathrm{sa}}$ using (11);
\STATE \hspace{0.5cm} \textbf{endfor} 
\STATE \hspace{0.5cm}obtain aggregated information $\bm{H}_{s}$ with spatial-attention weight at specific $s$ using (12);
\STATE \hspace{0.5cm} convert $\bm{H}_{s}$ into unified scale ${s}_{\mathrm{ms}}$ using (13);
\hspace{0.5cm} 
\STATE \textbf{endfor} 
\STATE combine all information $\left\{ \bm{H}_s \right\} _{s\in \mathcal{S}}$ to produce the ultimate aggregated semantic information $\bm{H}_{\mathrm{ms}}$ using (14);

\STATE create a temporary semantic information $\bm{H}_{\mathrm{h}}$ with historical prior information $\left.\{ \bm{F}_{i}^{t-k\mathrm{T}} \right\} _{k=\left\{1,2\cdots \mathrm{K} \right.\}}$ using (15);

\STATE get temporal-attention weight $\bm{W}^{\mathrm{ta}}$ using (16);
\STATE refine the final fused semantic information $\bm{H}_{i}$ with temporal-attention weight $\bm{W}^{\mathrm{ta}}$ using (17);

\STATE \textbf{Return} $\bm{H}_{i}$.
\end{algorithmic}
\label{alg2}
\end{algorithm}

\subsection{Semantic Information Aggregated from Spatial Dimension}
To enhance the  object feature representation from the spatial dimension, we first introduce a multi-scale attention module\cite{20} to extract spatial attention weights at various spatial resolutions and utilize the spatial-attention weight to indicate the importance of a collaborator's semantic information. Specifically, the spatial-attention weight of agent \emph{j} to agent \emph{i} at a specific spatial scale $s$ (${s\in \mathcal{S}}$)
is represented as $\bm{W}_{s,ji}^{\mathrm{sa}}$ and can be defined as 
\begin{align}
\bm{W}_{\!s,ji}^{\mathrm{sa}}\!=\mathrm{sa}\big( \mathrm{conv\!}\left.( \!\bm{F}_{i}^{ t}, s \right.) , \mathrm{conv}\!\left.( \!{\bm{F}_{j}^{ t-\tau _{ji}}}, s\right.)\big) ,
\end{align}
where $\mathcal{S}$ is the set of spatial scales, $\mathrm{conv}\left( F,s \right)$ denotes a convolution block \cite{39} that reshapes the scale of semantic information $F$ to $s$, and  $\mathrm{sa}\left( \cdot \right) $ represents the spatial attention operation, referring to the dot-product attention\cite{40}. Moreover, when \emph{j}  equals \emph{i}, the spatial-attention weight in Eq. (11) is reduced to the self-attention weight, resulting in a simpler and more consistent equation.

Next, we aggregate the semantic information of selected collaborators by assigning spatial-attention weights obtained in Eq. (11) to generate the aggregated semantic information $H_{s}$ with a specific scale $s$ as

\begin{align}
\bm{H}_{s}=\sum \nolimits_{j\in \mathcal{C}_{i}^{ t }\cup \left\{ i \right\}}{\bm{W}_{s,ji}^{\mathrm{sa}}\otimes \mathrm{conv}\!\left.( {\bm{F}_{j}^{ t-\tau _{ji}}},s \right.)}.
\end{align}

Subsequently, in order to ensure the scale consistency of the aggregated semantic information after spatial attention operation, we reshape it to a unified scale $\mathrm{s}_{\mathrm{ms}}$ by a deconvolution block \cite{41} as
\begin{align}
\bm{H}_{s}=\mathrm{deconv}\left.\!(\bm{H}_{s}, {s}_{\mathrm{ms}} \right.),
\end{align}
where $\mathrm{deconv}\left(\bm{H},s \right)$ denotes the deconvolution block that reshapes the scale of semantic information $\bm{H}$ to $s$.

Finally, we combine all the information $\left\{ \bm{H}_s \right\} _{s\in \mathcal{S}}$ acquired at various attention scales to produce the ultimate aggregated semantic information $\bm{H}_{\mathrm{ms}}$, that is,
\begin{align}
\bm{H}_{\mathrm{ms}}=\mathrm{concat}\left( \left\{ \bm{H}_s \right\} _{s\in \mathcal{S}} \right). 
\end{align}

The multi-scale attention module effectively learns the spatial correlation between the ego and its collaborators at different spatial resolutions, which enhances the aggregated features and presents a comprehensive representation of features. 

\subsection{Semantic Information Refined from Temporal Dimension}
Furthermore, considering that the historical semantic information of the ego contains rich object information, we design a historical prior short-term attention module to refine the aggregated semantic information from the temporal dimension. \par

First, the ego's  historical semantic information  $\left.\{ \bm{F}_{i}^{t-k\mathrm{T}} \right\} _{k=\left.\{1,2\cdots \mathrm{K}\right.\}}$ and the last-step aggregated semantic information $\bm{H}_{\mathrm{ms}}$  are concatenated to produce a new semantic information $\bm{H}_{\mathrm{h}}$, such as
\begin{align}
\bm{H}_{\mathrm{h}}=\mathrm{concat}\big( \bm{H}_{\mathrm{ms}}, \left.\{ \bm{F}_{i}^{t-k\mathrm{T}} \right\} _{k=\left\{ 1,2\cdots \mathrm{K} \right.\}} \big),
\end{align}
where $T$ represents the sensor sampling interval, and $K$ represents the number of historical frames.

Next, the semantic information $\bm{H}_{\mathrm{h}}$ is processed using average pooling and max pooling operations to generate average pooling features and max pooling features, respectively. Subsequently, we devise a short-term attention operation to process the two features to obtain the temporal-attention weight $\bm{W}^{\mathrm{ta}}$, that is,
\begin{align}
\bm{W}^{\mathrm{ta}}=\mathrm{ta}\big( \mathrm{avgpool}\left.( \bm{H}_{\mathrm{h}} \right.) , \mathrm{maxpool}\left.( \bm{H}_{\mathrm{h}} \right.) \big) ,
\end{align}
where $\mathrm{avgpool}\left( \cdot \right)$ and  $\mathrm{maxpool}\left( \cdot \right)$ represent the operations of average pooling  and  max pooling respectively, and $\mathrm{ta}\left( \cdot \right) $ denotes a temporal attention operation, and refers to the channel attention in \cite{18}.

Finally, the temporal-attention weight $\bm{W}^{\mathrm{ta}}$ is  multiplied by the semantic information $\bm{H}_{\mathrm{h}}$ to produce  the final fused semantic information $\bm{H}_{i}$ as
\begin{align}
\bm{H}_{i}= {\bm{W}^\mathrm{{ta}}  \otimes   \bm{H}_{\mathrm{h}} }.
\end{align}

By incorporating historical prior semantic information, the short-term attention module effectively captures the temporal correlation between past and present features. This enables it to identify discriminating object features, thereby enhancing the quality of object representation.\par

Overall, the proposed HPHA provides an enhanced and comprehensive feature representation for the fused semantic information $\bm{H}_{i}$, derived from both spatial and temporal dimensions. Consequently, the decoder $f_{\mathrm{dec}}\left( \cdot \right)$ can more efficiently decode the fused semantic information $\bm{H}_{i}$ to generate perception results $\widehat{Y}_{i}$, as depicted  in (6). \par

\section{Experimental evaluation}
This section provides a comprehensive performance comparison of the proposed Select2Col with the existing SOTA methods, such as V2VNet\cite{9}, V2X-Vit\cite{10}, and Where2comm\cite{11}. Furthermore, in order to demonstrate the superiority of Select2Col, we conduct various ablation studies to showcase the performance improvements. \par

\begin{table}[]
\centering
\caption{Experimental Parameters}
\renewcommand{\arraystretch}{1.1}
\begin{tabular}{|c|c|}
\hline
\multicolumn{1}{|c|}{\textbf{Parameter}} &  \multicolumn{1}{c|}{\textbf{Value}}  
 \\ \hline
Carrier frequency $f_{\mathrm{c}}$       & 5.9 Ghz                        \\
\hline
Total bandwidth       & 20 M                  \\
\hline
Transmit power $p_{ji}^{\mathrm{tx}}$       & 23 dbm                          \\
\hline
Power of noise $p_{ji}^{\mathrm{noise}}$             &-95 dbm $\sim$  -110 dbm                                   \\
\hline
Sensor asynchronous overhead  $t_{ji}^{\mathrm{asyn}}$        &-100 ms $\sim$  100 ms     \\
\hline
Semantic  extraction time $t_{ji}^\mathrm{{ext}}$       &20 ms $\sim$ 40 ms   \\
\hline
Multi-scale attention scale set $\mathcal{S}$   & \begin{tabular}[c]{@{}c@{}}{[}64, 96, 352{]}\\ {[}128, 48, 176{]}\\ {[}256, 24, 88{]}\end{tabular} \\ 
\hline
Multi-scale attention output scale $\mathrm{s}_{\mathrm{ms}}$ &{[}128, 96, 352{]} \\ 
\hline
Sensor sampling interval $T$      & 100 ms                              \\ 
\hline
Number of historical prior frames $K$   & 2\\
\hline
\end{tabular}
\end{table}

\subsection{Experimental Settings}
Rather than evaluating on popular datasets such as KITTI\cite{42} and Nuscenes\cite{43}, which mainly provide single-agent samples, all experiments in this article are conducted on open collaborative perception datasets OPV2V\cite{14}, V2XSet\cite{10} and V2V4Real\cite{22}. 
Specifically, OPV2V is the first large-scale open dataset targeted at vehicle-to-vehicle collaborative perception, generated using CARLA\cite{44} and SUMO\cite{45}. On the other hand, V2XSet is  an open dataset that focuses on vehicle-to-vehicle and vehicle-to-infrastructure collaborative perception, generated employing CARLA and OpenCDA\cite{46}. Consequently, these two popular datasets are inherently distinct.
In addition, the V2V4Real dataset which comprises collaborative perception data extracted from real-world road scenarios spans a three-day period of driving in Columbus, Ohio, covering a total driving area of $410$ kilometers, incorporating $347$ kilometers of highway roads and $63$ kilometers of city roads.\par

To comprehensively evaluate the performance of our proposed Select2Col, we utilize the average accuracy (AP) at IoU (Intersection over Union) thresholds of 0.3, 0.5, and 0.7.  All experiments are conducted on an X86 station equipped with an Intel TM i7-11700 @2.50 GHz, 128-core CPU, 256 GB RAM, and NVIDIA RTX3090 GPU.\par

As presented in TABLE III, the key parameters employed in our study are in line with V2VNet\cite{9}, V2X-Vit\cite{10}, and Where2comm\cite{11}. In addition, Eq. (3) is further adopted for a more realistic calculation of the transmission time, where the related parameters are consistent with the literature \cite{47}. Notably, we allocate equal bandwidth to each individual agent. Consequently, our setting is more precise compared to V2X-Vit\cite{10} and Where2comm\cite{11}, which employ a fixed transmission rate of $27$ Mbps for all agents. Furthermore, the semantic information extraction latency $t_{ji}^{\mathrm{ext}}$ is obtained based on our computing device. More experimental parameters can be found in our open-source code. \par

\subsection{Perception Performance Comparison}
In this experiment, we evaluate the overall perception performance of our proposed Select2Col on a total of $664$ test data from the OPV2V dataset, $1,466$ test samples from the V2XSet dataset, and $1,993$ test samples from the V2V4Real dataset. To reduce randomness, we use their average inference results as our experimental outcomes.\par

\begin{table}
\centering
\caption{Overall Perception Performance}
\renewcommand{\arraystretch}{1.1}
\begin{tabular}{|c|c|ccc|}
\hline
\multirow{2}{*}{\makebox[1.2cm]{Dataset}} & \multirow{2}{*}{\makebox[1.5cm]{Methods}} & \multicolumn{3}{c|}{\makebox[3cm]{AP (\%)}}                                             \\ \cline{3-5} 
                         &                          & \multicolumn{1}{c|}{\makebox[1.0cm]{IoU 0.3}}          & \multicolumn{1}{c|}{\makebox[1.0cm]{IoU 0.5}}          & \makebox[1.0cm]{IoU 0.7}                   \\ \hline
\multirow{5}{*}{\makebox[1.2cm]{OPV2V}}   
                         & \makebox[1.5cm]{No fusion}           & \multicolumn{1}{c|}{\makebox[1.0cm]{79.81}}          & \multicolumn{1}{c|}{\makebox[1.0cm]{77.70}}          & \makebox[1.0cm]{62.12}                  \\ \cline{2-5} 
                         &\makebox[1.5cm]{V2X-Vit}                  & \multicolumn{1}{c|}{\makebox[1.0cm]{87.09}}          & \multicolumn{1}{c|}{\makebox[1.0cm]{85.89}}          & \makebox[1.0cm]{75.56}           \\ \cline{2-5} 
                         &\makebox[1.5cm]{V2VNet}                   & \multicolumn{1}{c|}{\makebox[1.0cm]{82.71}}          & \multicolumn{1}{c|}{\makebox[1.0cm]{80.38}}          & \makebox[1.0cm]{52.96}             \\ \cline{2-5} 
                         &\makebox[1.5cm]{Where2comm}               & \multicolumn{1}{c|}{\makebox[1.0cm]{86.71}}          & \multicolumn{1}{c|}{\makebox[1.0cm]{85.20}}          & \makebox[1.0cm]{71.35}            \\ \cline{2-5} 
                         & \textbf{\makebox[1.5cm]{Select2Col}}         & \multicolumn{1}{c|}{\textbf{\makebox[1.0cm]{89.80}}} & \multicolumn{1}{c|}{\textbf{\makebox[1.0cm]{88.51}}} & \textbf{\makebox[1.0cm]{77.65}} \\ \hline
\multirow{5}{*}{\makebox[1.2cm]{V2XSet}}  
                         & \makebox[1.5cm]{No fusion}           & \multicolumn{1}{c|}{\makebox[1.0cm]{75.55}}          & \multicolumn{1}{c|}{\makebox[1.0cm]{71.17}}          & \makebox[1.0cm]{47.12}             \\ \cline{2-5} 
                         &\makebox[1.5cm]{V2X-Vit}                  & \multicolumn{1}{c|}{\makebox[1.0cm]{81.54}}          & \multicolumn{1}{c|}{\makebox[1.0cm]{78.76}}          & \makebox[1.0cm]{62.36}               \\ \cline{2-5} 
                         &\makebox[1.5cm]{V2VNet}                   & \multicolumn{1}{c|}{\makebox[1.0cm]{77.85}}          & \multicolumn{1}{c|}{\makebox[1.0cm]{72.27}}          & \makebox[1.0cm]{46.99}              \\ \cline{2-5} 
                         &\makebox[1.5cm]{Where2comm}               & \multicolumn{1}{c|}{\makebox[1.0cm]{82.70}}          & \multicolumn{1}{c|}{\makebox[1.0cm]{79.05}}          & \makebox[1.0cm]{57.16}           \\ \cline{2-5} 
                         &\textbf{\makebox[1.5cm]{Select2Col}}         & \multicolumn{1}{c|}{\textbf{\makebox[1.0cm]{87.18}}} & \multicolumn{1}{c|}{\textbf{\makebox[1.0cm]{84.25}}} & \textbf{\makebox[1.0cm]{64.92}}  \\ \hline
\multirow{4}{*}{\makebox[1.2cm]{V2V4Real}}  
                         & \makebox[1.5cm]{No fusion}           & \multicolumn{1}{c|}{\makebox[1.0cm]{46.77}}          & \multicolumn{1}{c|}{\makebox[1.0cm]{37.23}}          & \makebox[1.0cm]{15.55}             \\ \cline{2-5} 
                         &\makebox[1.5cm]{V2X-Vit}                 & \multicolumn{1}{c|}{\makebox[1.0cm]{55.08}}          & \multicolumn{1}{c|}{\makebox[1.0cm]{47.78}}          & \makebox[1.0cm]{23.74}               \\ \cline{2-5} 
                         &\makebox[1.5cm]{Where2comm}               & \multicolumn{1}{c|}{\makebox[1.0cm]{60.28}}          & \multicolumn{1}{c|}{\makebox[1.0cm]{48.32}}          & \makebox[1.0cm]{20.55}           \\ \cline{2-5} 
                         &\textbf{\makebox[1.5cm]{Select2Col}}         & \multicolumn{1}{c|}{\textbf{\makebox[1.0cm]{62.20}}} & \multicolumn{1}{c|}{\textbf{\makebox[1.0cm]{50.19}}} & \textbf{\makebox[1.0cm]{24.89}}  \\ \hline
\end{tabular}
\end{table}

As presented in TABLE IV, our proposed Select2Col outperforms V2VNet\cite{9}, V2X-Vit\cite{10}, and Where2comm\cite{11} in terms of AP. For example, at an IoU threshold of 0.7, Select2Col achieves gains of $24.69$\%, $2.09$\%, and $6.30$\% in AP performance compared to V2VNet, V2X-Vit, and Where2comm, respectively, on the OPV2V dataset.
Furthermore, Select2Col proves to be highly effective and robust, even in real-world situations obtained from V2V4Real. Notably, at an IoU threshold of 0.3, Select2Col demonstrates a remarkable improvement of $15.43$\% in AP performance compared to single-agent perception (i.e., no fusion).

\textit{Discussions}: The remarkable results achieved by Select2Col are attributed to its introduction of innovative features. First, selecting contributive collaborators while pruning less effective ones can reduce noise and interference, thereby improving the perception performance. Second, by enhancing the effective features while suppressing noise, the effectiveness of semantic information is boosted. Third, a multi-scale attention module aggregates semantic information from different collaborators with IoSI-consistent weights from the spatial dimension, which provides a comprehensive representation of features. Finally, a short-term attention module is applied with historical prior semantic information, which refines the final fused semantic information from the temporal dimension and further improves the perception performance. Overall, unlike other works that only focus on the spatial dimension, Select2Col comprehensively considers collaborative perception from both spatial and temporal dimensions and yields  unparalleled performance.\par

\subsection{Computational Efficiency Comparison}

Table V and Fig. 5 present the computing efficiency of Select2Col. Specifically, as demonstrated in TABLE V, our Select2Col has fewer network model parameters than V2X-Vit and V2VNet and a similar number of parameters to Where2comm. In other words, compared to other spatial dimension-only solutions, the incorporation of  both spatial and temporal dimensions of IoSI in selecting collaborators and fusing semantic information does not add any computational complexity. Furthermore, as depicted in Fig. 5, the inference time of our proposed Select2Col is significantly lower than that of V2VNet and V2X-Vit, and slightly higher than that of Where2comm due to the introduction of the lightweight GNN and the short-term attention module; however, Select2Col outperforms Where2comm in terms of perception performance. The inference time of Select2Col is much less than $100$ ms, which is  acceptable  given  the perception cycle (i.e., $10$ Hz) of the LiDAR sensor device.\par

\textit{Discussions}: Compared to Where2comm, the newly added lightweight GNN and short-term attention modules can enhance perception performance and are easily handled by mainstream computing devices without an excessive computational burden. Overall, Select2Col generates superior perception performance with a high computational efficiency.\par

\begin{figure}[!t]
\centering
\includegraphics[width=0.45\textwidth]{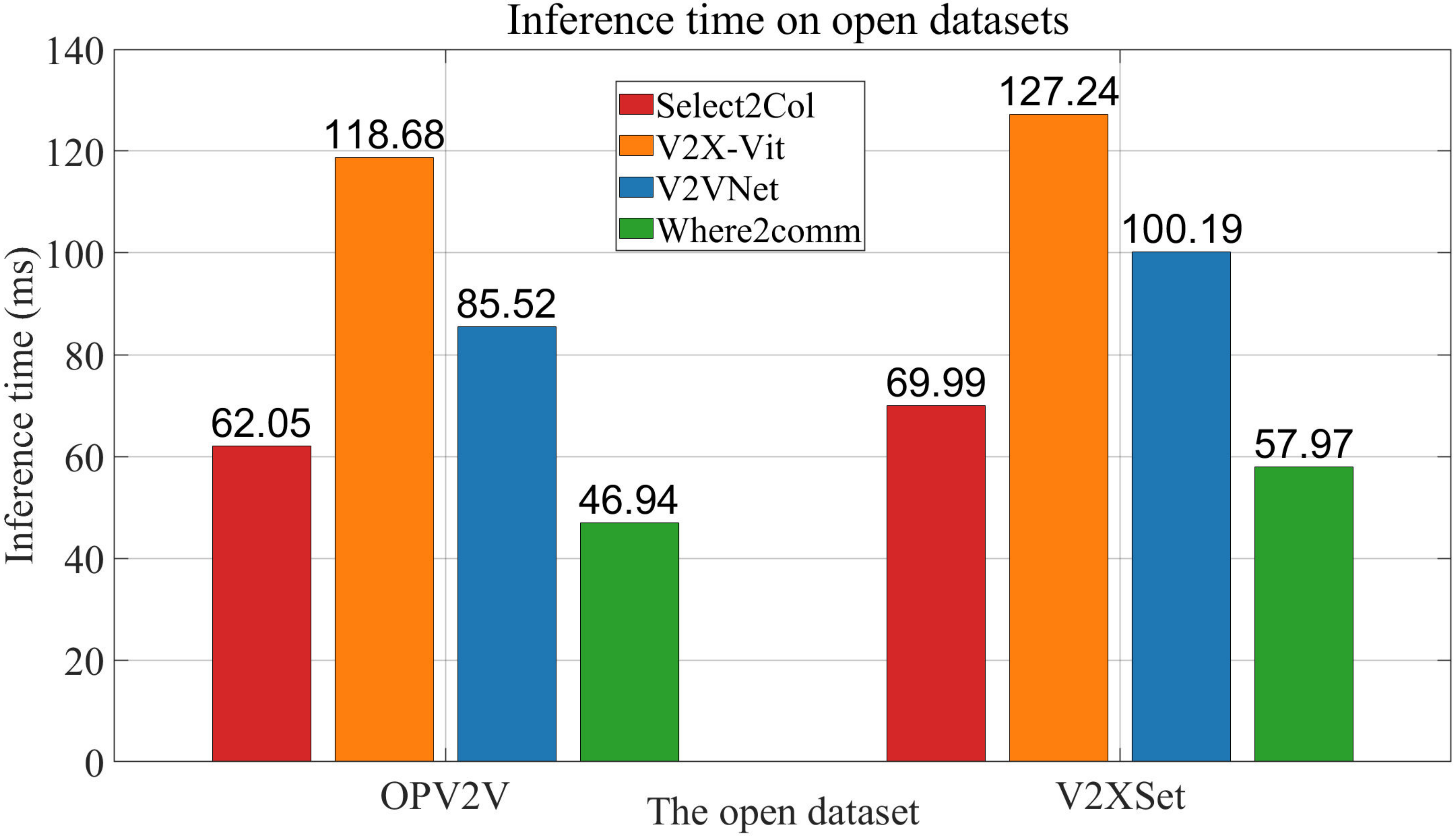}
\caption{The comparison of computing efficiency.} 
\label{Fig.5}
\end{figure}

\begin{table}[!t]
\centering
\caption{Number of  network model parameters}
\renewcommand{\arraystretch}{1.1}
\begin{tabular}{|c|c|}
\hline
\textbf{Works} & \begin{tabular}[c]{@{}c@{}}\textbf{Number of} \\  \textbf {paramters  (M)}\end{tabular}    \\ \hline
V2VNet         &14.6122       \\ \hline
V2X-Vit        &12.4554    \\ \hline
Where2comm     &8.0574     \\ \hline
\textbf{Select2Col}        &\textbf{8.2875}   \\ \hline
\end{tabular}
\end{table}

\subsection{Perception Performance under Localization Noise}
Notably, localization noise can be widely observed due to the impact of spatial and temporal misalignment. To evaluate the robustness of our proposed Select2Col, we add localization noise to the test data, which follows a Gaussian distribution with a standard deviation of 0.2 for both position and heading.\par

As presented in TABLE  VI, localization noise impairs the perception performance of all approaches, and the reduction is more noticeable when the IoU threshold is larger (i.e., the detection difficulty is increased). This indicates that high-precision detection is more vulnerable to noise. However, compared to other methods, our proposed Select2Col shows superior noise suppression ability. For instance, on the V2XSet dataset, Select2Col maintains an AP of $78.74$\% and $58.41$\% at IoU thresholds of 0.5 and 0.7, presenting an improvement of $10.20$\%/$22.18$\%, $1.72$\%/$1.51$\% and $3.85$\%/$7.26$\% compared to V2VNet, V2X-Vit, and Where2comm, respectively. The same conclusion can be observed in the OPV2V dataset.\par

\textit{Discussions}: Select2Col introduces historical prior semantic information and utilizes a short-term attention module to learn the correlation of semantic information from the temporal dimension. This enables the suppression of spatial localization noise for the current semantic information by using historical information. As a result, the localization noise has less impact on Select2Col.\par

\begin{table}
\centering
\caption{Perception Performance under noise}
\renewcommand{\arraystretch}{1.1}
\begin{tabular}{|c|c|ccc|}
\hline
\multirow{2}{*}{\makebox[1.2cm]{Dataset}} & \multirow{2}{*}{\makebox[1.5cm]{Methods}} & \multicolumn{3}{c|}{\makebox[3cm]{AP (\%)}}                                             \\ \cline{3-5} 
                         &                          & \multicolumn{1}{c|}{\makebox[1.0cm]{IoU 0.3}}          & \multicolumn{1}{c|}{\makebox[1.0cm]{IoU 0.5}}          & \makebox[1.0cm]{IoU 0.7}                   \\ \hline
\multirow{5}{*}{\makebox[1.2cm]{OPV2V}}   
                         & \makebox[1.5cm]{No fusion}           & \multicolumn{1}{c|}{\makebox[1.0cm]{79.81}}          & \multicolumn{1}{c|}{\makebox[1.0cm]{77.70}}          & \makebox[1.0cm]{62.12}                  \\ \cline{2-5} 
                         &\makebox[1.5cm]{V2X-Vit}                  & \multicolumn{1}{c|}{\makebox[1.0cm]{86.76}}          & \multicolumn{1}{c|}{\makebox[1.0cm]{85.41}}          & \makebox[1.0cm]{74.51}           \\ \cline{2-5} 
                         &\makebox[1.5cm]{V2VNet}                   & \multicolumn{1}{c|}{\makebox[1.0cm]{82.24}}          & \multicolumn{1}{c|}{\makebox[1.0cm]{78.33}}          & \makebox[1.0cm]{44.72}             \\ \cline{2-5} 
                         &\makebox[1.5cm]{Where2comm}               & \multicolumn{1}{c|}{\makebox[1.0cm]{86.47}}          & \multicolumn{1}{c|}{\makebox[1.0cm]{84.30}}          & \makebox[1.0cm]{66.52}            \\ \cline{2-5} 
                         & \textbf{\makebox[1.5cm]{Select2Col}}         & \multicolumn{1}{c|}{\textbf{\makebox[1.0cm]{89.73}}} & \multicolumn{1}{c|}{\textbf{\makebox[1.0cm]{88.20}}} & \textbf{\makebox[1.0cm]{74.54}} \\ \hline
\multirow{5}{*}{\makebox[1.2cm]{V2XSet}}  
                         & \makebox[1.5cm]{No fusion}           & \multicolumn{1}{c|}{\makebox[1.0cm]{75.55}}          & \multicolumn{1}{c|}{\makebox[1.0cm]{71.17}}          & \makebox[1.0cm]{47.12}             \\ \cline{2-5} 
                         &\makebox[1.5cm]{V2X-Vit}                  & \multicolumn{1}{c|}{\makebox[1.0cm]{80.77}}          & \multicolumn{1}{c|}{\makebox[1.0cm]{77.02}}          & \makebox[1.0cm]{56.90}               \\ \cline{2-5} 
                         &\makebox[1.5cm]{V2VNet}                   & \multicolumn{1}{c|}{\makebox[1.0cm]{76.12}}          & \multicolumn{1}{c|}{\makebox[1.0cm]{68.54}}          & \makebox[1.0cm]{36.23}              \\ \cline{2-5} 
                         &\makebox[1.5cm]{Where2comm}               & \multicolumn{1}{c|}{\makebox[1.0cm]{80.74}}          & \multicolumn{1}{c|}{\makebox[1.0cm]{74.89}}          & \makebox[1.0cm]{51.15}           \\ \cline{2-5} 
                         &\textbf{\makebox[1.5cm]{Select2Col}}         & \multicolumn{1}{c|}{\textbf{\makebox[1.0cm]{84.90}}} & \multicolumn{1}{c|}{\textbf{\makebox[1.0cm]{78.74}}} & \textbf{\makebox[1.0cm]{58.41}}  \\ \hline
\end{tabular}
\end{table}

\subsection{Perception Performance under Distance}

Collaborative perception improves the agent's ability to perceive objects that are located far away. Given the scarcity of object samples beyond $120$ meters in test datasets, we solely evaluate the perception performance of objects within $120$ meters to avoid the inherent randomness arising from an inadequate number of samples. Fig. 6 to Fig. 8 illustrate the perception performance of our proposed Select2Col and other SOTA methods at different object distances on the OPV2V and V2XSet datasets.\par

Intuitively, as the distance between the agent and the object increases, the probability of accurately detecting the object decreases.  Consequently, the perception performance tends to decline as the distance increases. Nonetheless, our proposed Select2Col is still able to achieve the highest perception accuracy in long-distance perception scenarios. For instance, at an IoU threshold of 0.5 and object distance of 60 meters, Select2Col improves the AP performance by $16.97$\%/$9.18$\%/$10.07$\% in comparison to V2VNet, V2X-Vit and Where2comm, respectively, on the V2XSet dataset.\par

\textit{Discussions}: The efficacy of Select2Col in perceiving distant objects lies in its ability to enhance the fused semantic information of collaborators in both spatial and temporal dimensions with multi-scale attention and short-term attention, thus enhancing the capability to perceive distant objects.  \par

\begin{figure}[!t]
\centering
\includegraphics[width=0.48\textwidth]{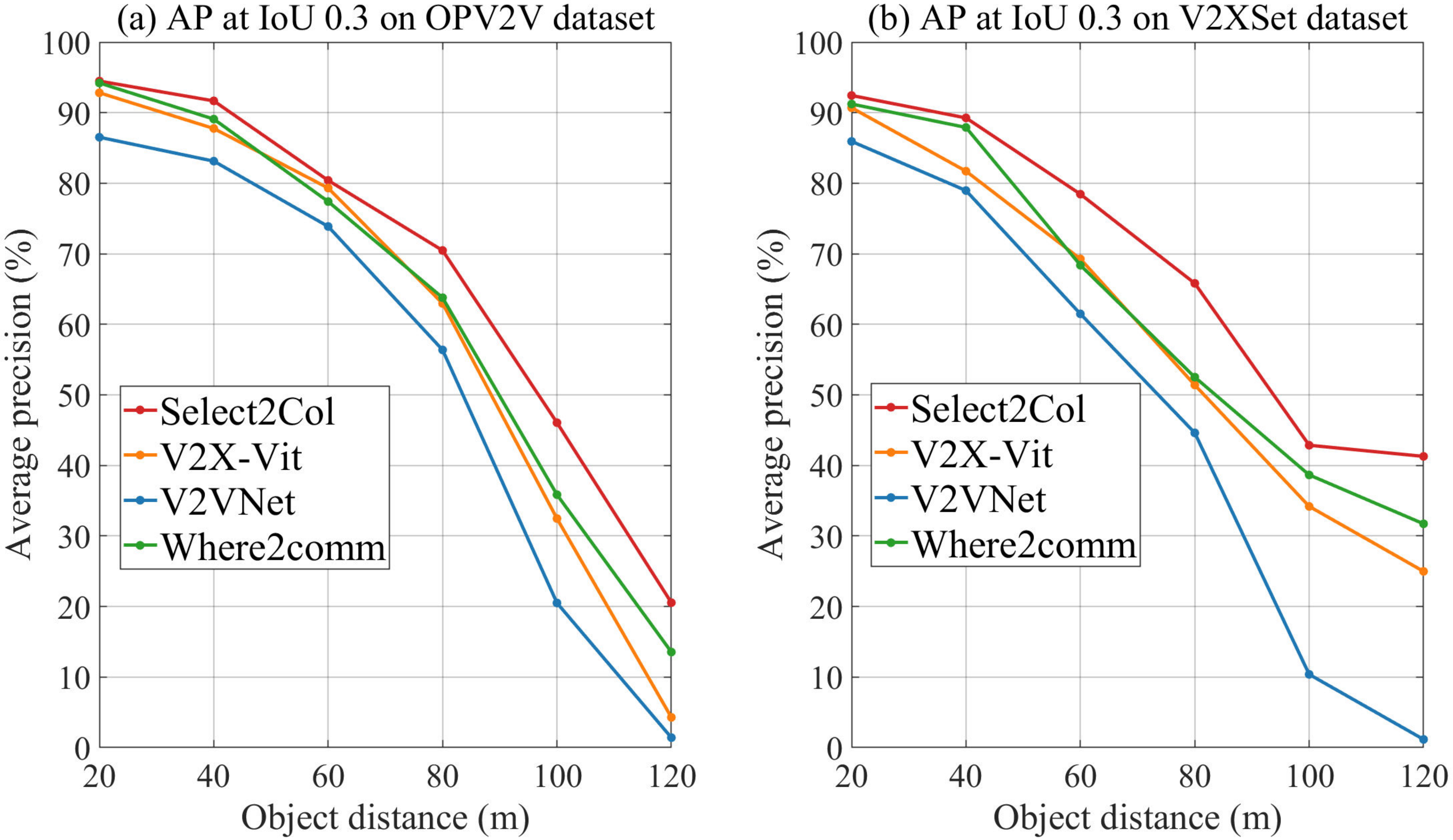}
\caption{Perception performance under distance at IoU 0.3.} 
\label{Fig.6}
\end{figure}

\begin{figure}[!t]
\centering
\includegraphics[width=0.48\textwidth]{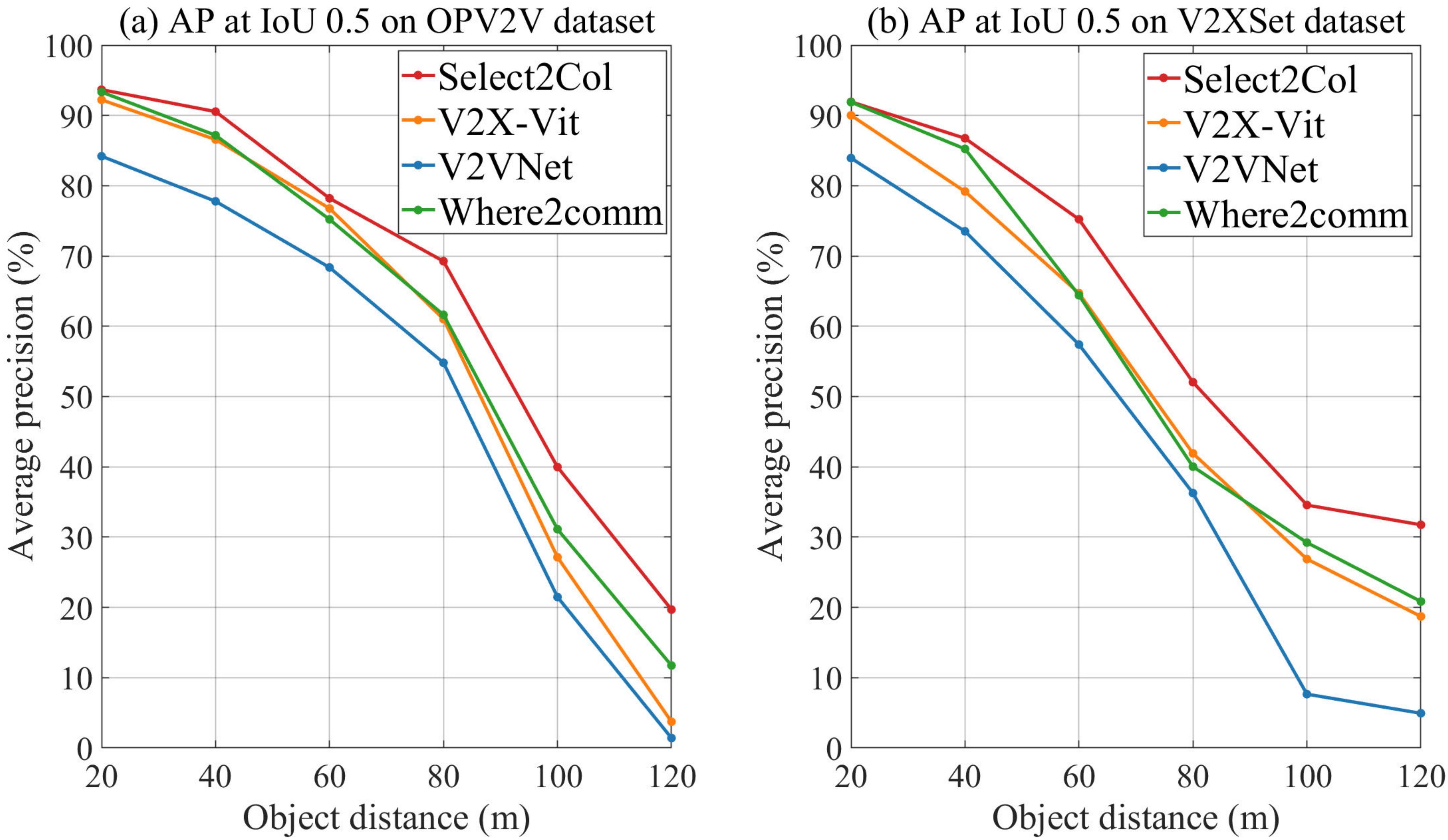}
\caption{Perception performance under distance at IoU 0.5.} 
\label{Fig.7}
\end{figure}

\begin{figure}[!t]
\centering
\includegraphics[width=0.48\textwidth]{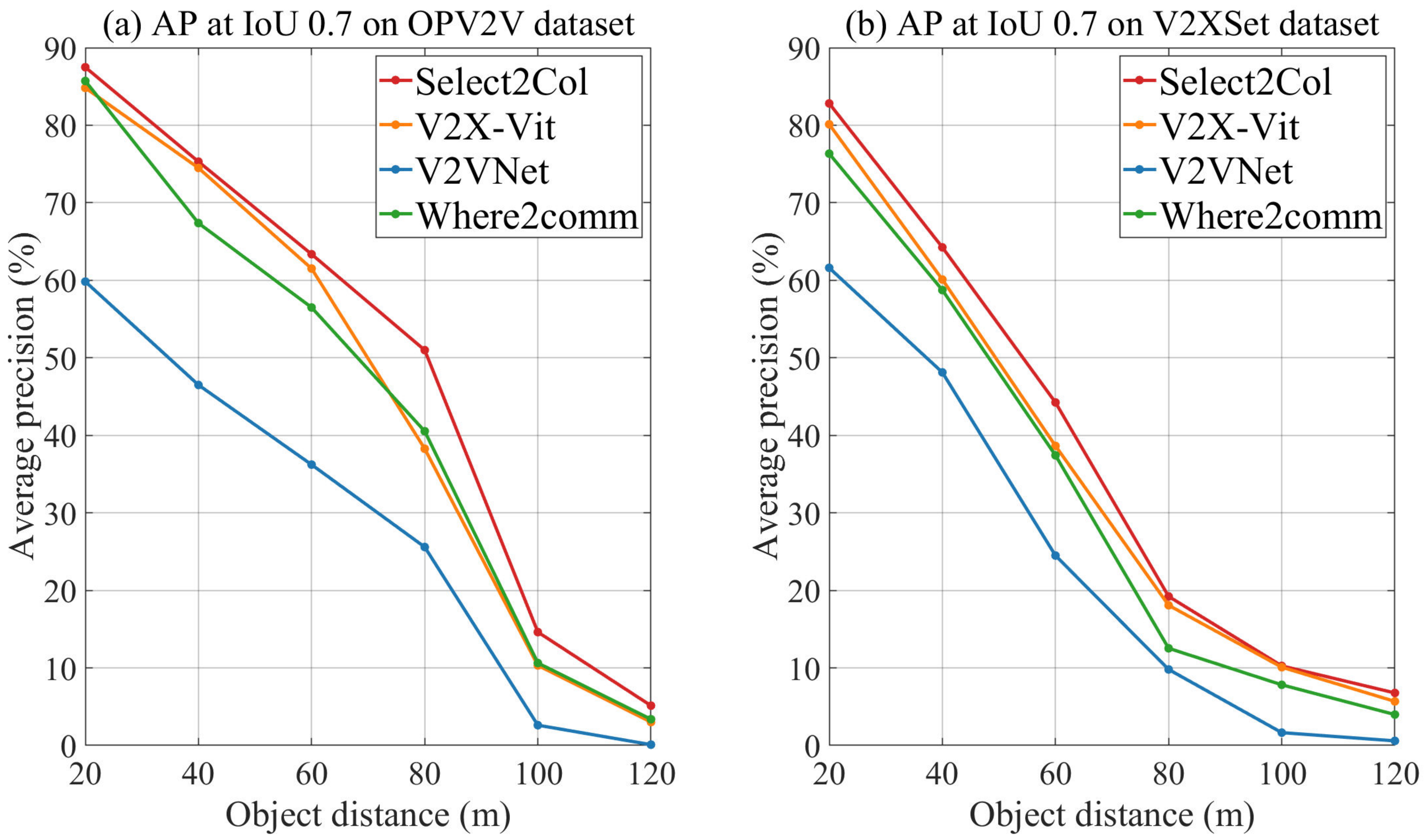}
\caption{Perception performance under distance at IoU 0.7.} 
\label{Fig.8}
\end{figure}

\begin{figure}[!t]
\centering
\includegraphics[width=0.48\textwidth]{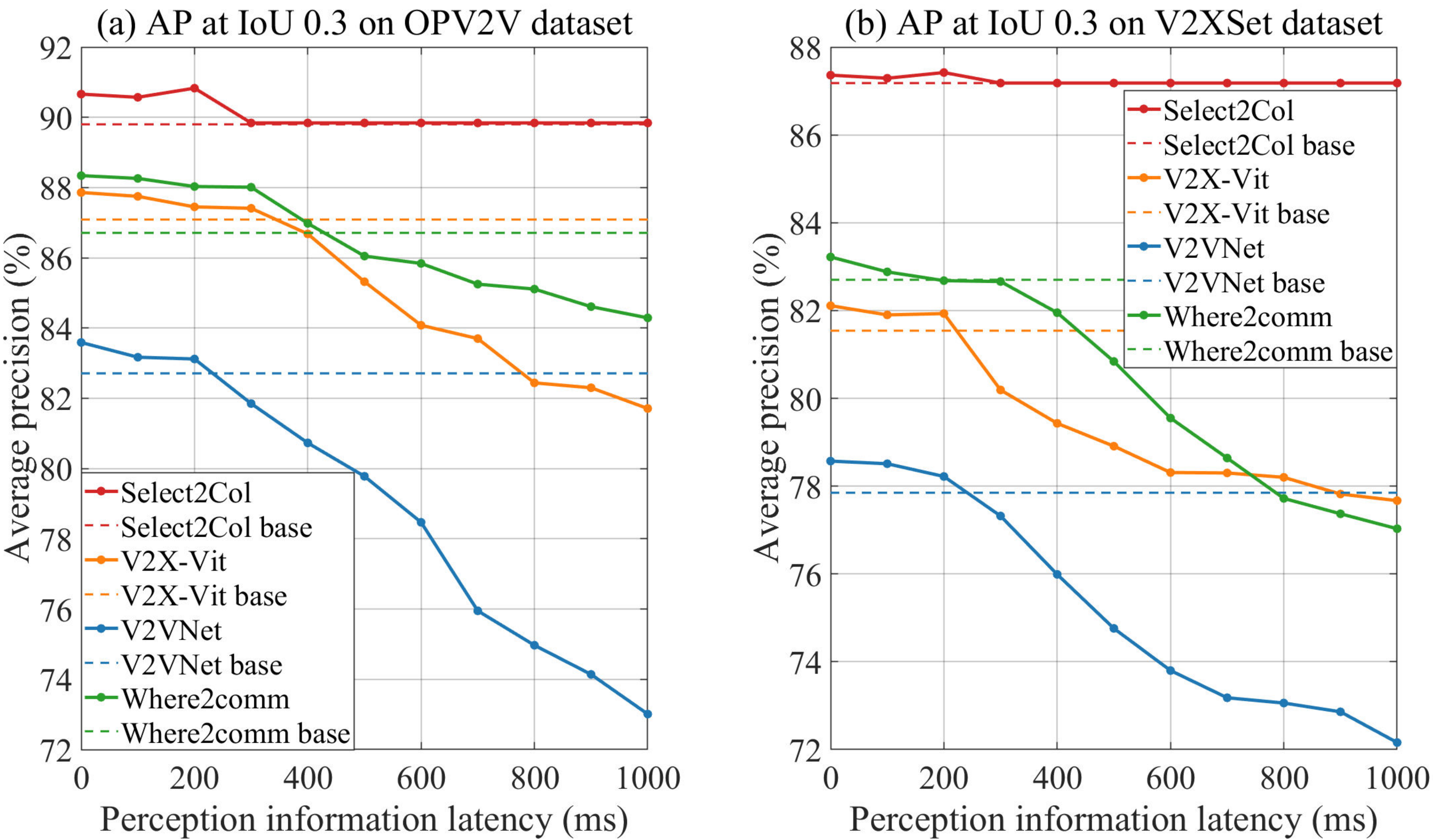}
\caption{Perception performance under latency at IoU 0.3.} 
\label{Fig.9}
\end{figure}

\begin{figure}[!t]
\centering
\includegraphics[width=0.48\textwidth]{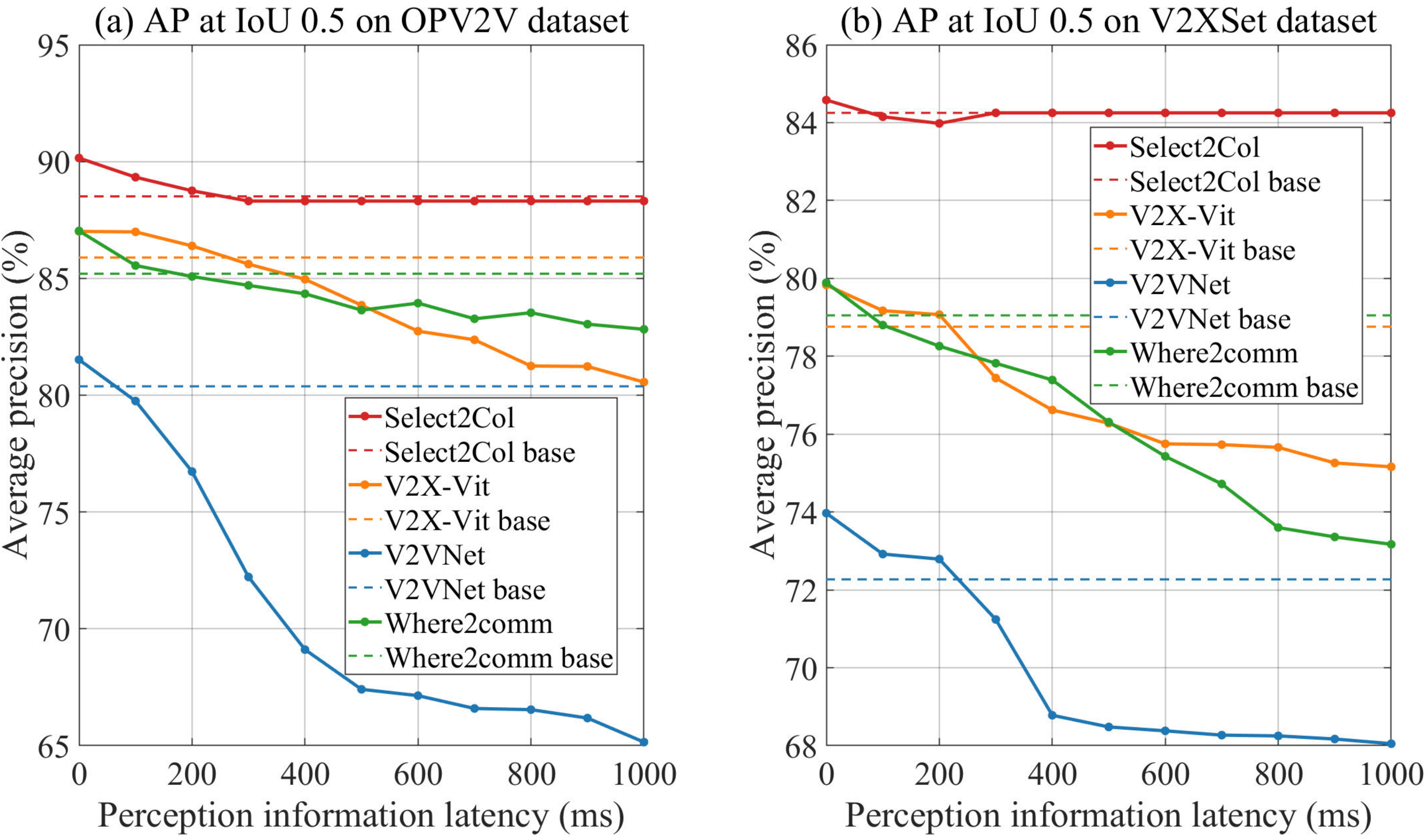}
\caption{Perception performance under latency at IoU 0.5.} 
\label{Fig.10}
\end{figure}

\begin{figure}[!t]
\centering
\includegraphics[width=0.48\textwidth]{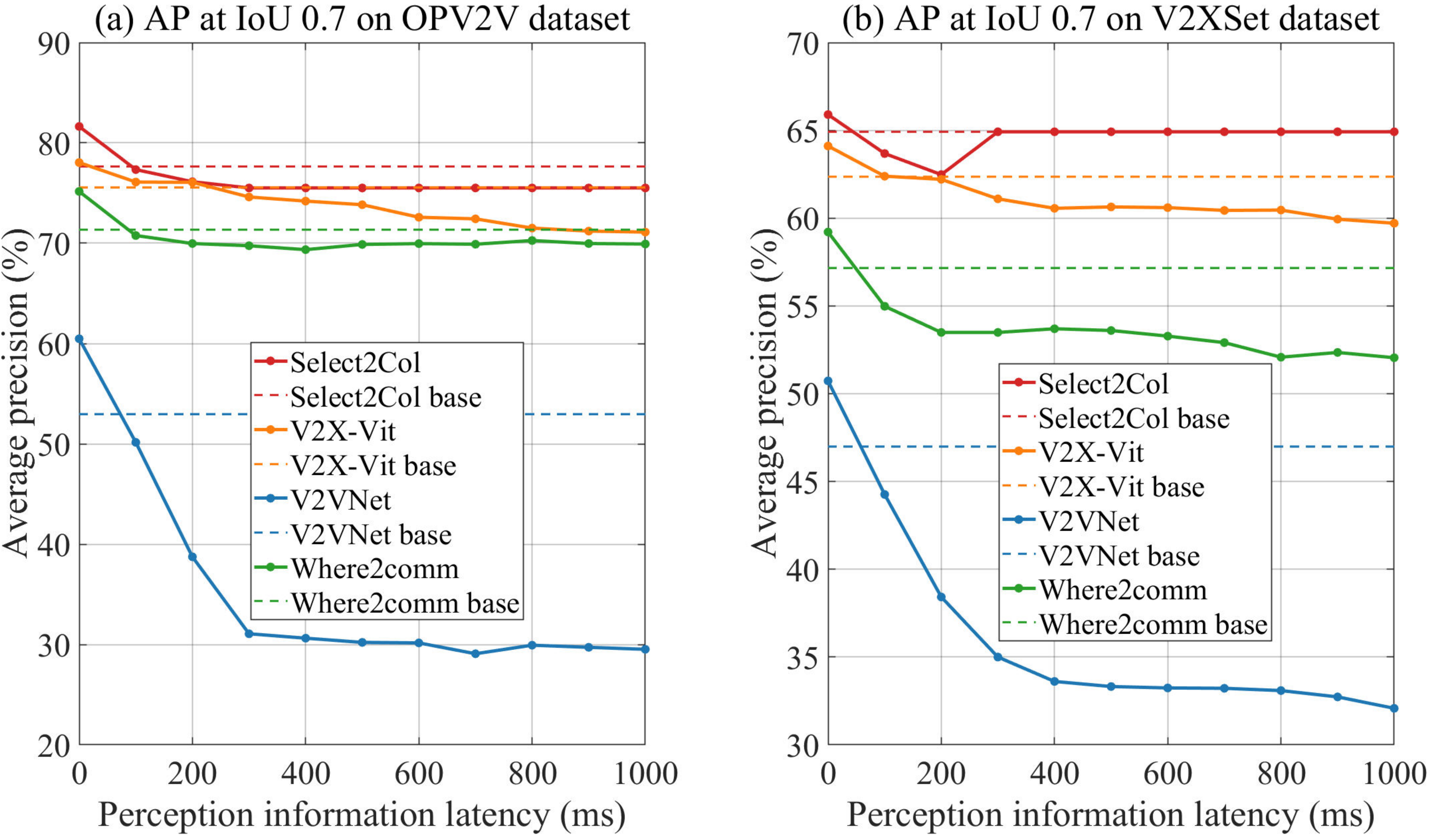}
\caption{Perception performance under latency at IoU 0.7.} 
\label{Fig.11}
\end{figure}

\subsection{Perception Performance under Latency}
This experiment aims to evaluate the influence of information latency on perception performance. To achieve this, we randomly add a collaborator with a specific latency in test samples.\par

Fig. 9 to Fig. 11 demonstrate the corresponding results under different IoU thresholds. As depicted in these figures, the collaborator with a small latency has the potential to enhance perception performance, whereas a large latency has a negative impact. Furthermore, our proposed Select2Col exhibits the best performance among all approaches and remains performance-stable even under large latency. For instance, at an IoU threshold of 0.3, when the latency of the collaborator is $400$ ms, Select2Col enhances the AP performance by $11.19$\%/$7.75$\%/$5.23$\% on the V2XSet dataset compared to V2VNet, V2X-Vit, and Where2comm, respectively.\par

\textit{Discussions}:  As anticipated, the results of the experiment support our hypothesis that the collaborator with notable latency has a minimal or even negative impact on enhancing the ego's perception performance. Furthermore, it validates  the significance of collaborator selection and strongly supports the effectiveness of selecting collaborators from both temporal and spatial dimensions to eliminate the negative impact. In Select2col, we utilize a collaborator selection method that considers both temporal and spatial dimensions, thus effectively eliminating collaborators with significant latency. In contrast, other methods either neglect collaborator selection entirely or solely rely on spatial criteria, resulting in a decline in perception performance as latency increases. Consequently, Select2Col ensures stable perception performance even in situations with notable latency and is more robust than other methods.\par

\subsection{Ablation Studies}
In this subsection, we evaluate the effectiveness of the innovations incorporated in our proposed Select2Col. As previously mentioned, Select2Col includes two novel innovations: i) the collaborator selection method; ii) the  semantic information fusion algorithm HPHA. \par

\begin{table}
\centering
\caption{Innovative Component Ablation Study Results}
\renewcommand{\arraystretch}{1.2}
\begin{tabular}{|m{0.8cm}<{\centering}|m{0.7cm}<{\centering}|m{1.4cm}<{\centering}|m{0.8cm}<{\centering}|m{0.8cm}<{\centering}|m{0.8cm}<{\centering}|}
\hline
\textbf{Open dataset} & \textbf{HPHA} &\textbf{Collaborator selection}  &\textbf{AP at IoU0.3} &\textbf{AP at IoU0.5} &\textbf{AP at IoU0.7}\\
\hline
\multirow{3}* {OPV2V} &\ding{56} &\ding{56} &79.81	&77.70	&62.12 \\
\cline{2-6}
                      &\ding{52} &\ding{56} &88.11	&86.13	&73.91 \\
\cline{2-6}
			      &\ding{52} &\ding{52} &\textbf{89.80}	&\textbf{88.51}	&\textbf{77.65} \\
\cline{2-6}
\hline

\multirow{3}* {V2XSet}&\ding{56} &\ding{56} &75.55	&71.17	&47.12 \\
\cline{2-6}
                      &\ding{52} &\ding{56} &85.47	&81.85	&61.52 \\
\cline{2-6}
			      &\ding{52} &\ding{52} &\textbf{87.18}	&\textbf{84.25}	&\textbf{64.92} \\
\cline{2-6}

\hline
\end{tabular}
\end{table}

TABLE VII presents the results of our ablation studies. The outcome clearly illustrates that each innovative component significantly contributes to the advancement of perception performance. Specifically, the collaborator selection method enhances the perception performance of AP by $1.71$\%, $2.40$\%, and $3.40$\%, respectively, on the V2XSet dataset. Moreover, HPHA improves the perception performance of AP by $9.92$\%, $10.68$\%, and $14.4$\%  on the V2XSet dataset. In addition, we conclude that the gains from the collaborator selection method and HPHA are more pronounced when the value of the IoU threshold is larger (i.e., the detection difficulty is increased). This trend is in line with our previous experimental findings and is more apparent in HPHA. \par

\textit{Discussions}: The collaborator selection method benefits the removal of unsuitable collaborators that cause noise, resulting in improved perception performance. Likewise, the gain from HPHA is obtained due to the efficient utilization of the correlation between temporal and spatial dimensions embedded in the semantic information for information fusion. Thus, both techniques can effectively enhance the perception performance.\par

\section{CONCLUSION}
In this article, we have proposed Select2Col, a novel collaborative perception framework that improves perception performance based on IoSI from both spatial and temporal dimensions. Specifically, we have designed a collaborator selection method that capably selects contributive collaborators efficiently. To further boost the perception performance, we have presented a semantic information fusion algorithm named HPHA by integrating a multi-scale attention module and a short-term attention module to capture the IoSI from both spatial and temporal dimensions and aggregate semantic information by assigning IoSI-consistent weights. Extensive experimental results are conducted on three open datasets, OPV2V, V2XSet, and V2V4Real, and demonstrate that Select2Col outperforms SOTA methods in terms of perception performance.\par

There still exist many promising means to improve the performance of our Select2Col framework. For instance, the current approach for calculating enhanced weights overlooks the IoU threshold, leading to situations where a collaborator may be deemed effective at a low IoU threshold but fails to enhance perception performance at higher IoU thresholds. Thus, a more accurate enhanced weight determination method is highly anticipated. Furthermore, the sharing of perception information inherently carries the risk of privacy leakage. Therefore, it is worthwhile to investigate collaborative privacy protection techniques with malicious behavior avoidance to enhance collaborators' willingness to participate in sharing information.

\par

\section*{Acknowledgments}
This work was supported in part by the  National Key Research and Development Program of China under Grant 2021YFB2900200, in part by the National Natural Science Foundation of China under Grant 62071425, in part by the Zhejiang Key Research and Development Plan under Grant 2022C01129, and in part by the Zhejiang Provincial Natural Science Foundation of China under Grant LR23F010005.

\bibliographystyle{IEEEtran} 
\bibliography{ref}

%
%




\newpage

\vfill

\end{document}